\documentclass[11pt]{article}

\makeatletter
\def\input@path{{./}{../}}
\makeatother

\usepackage[margin=1in]{geometry}
\usepackage[T1]{fontenc}
\usepackage{amsmath,amssymb,amsthm}
\usepackage{booktabs}
\usepackage{graphicx}
\graphicspath{{./}{../}{../figures/}}   %
\usepackage{float}
\usepackage{multirow}
\usepackage[font=small,labelfont=bf,labelsep=period,skip=6pt]{caption}
\usepackage{xcolor}
\usepackage[colorlinks=true,linkcolor=blue,citecolor=blue,urlcolor=blue]{hyperref}
\usepackage[round]{natbib}

\newcounter{algocounter}
\newcommand{\algocaption}[1]{%
  \refstepcounter{algocounter}%
  \par\noindent\textbf{Algorithm~\thealgocounter.} #1\par\smallskip}
\newenvironment{myalg}{%
  \begin{figure}[tbp]\centering
  \begin{minipage}{0.95\linewidth}
  \hrule height 0.8pt \vspace{4pt}
}{%
  \vspace{4pt}\hrule height 0.8pt
  \end{minipage}
  \end{figure}
}

\makeatletter
\newsavebox{\arx@rbox}
\let\arx@origresizebox\resizebox
\renewcommand{\resizebox}[3]{%
  \sbox{\arx@rbox}{#3}%
  \ifdim\wd\arx@rbox>#1\relax\arx@origresizebox{#1}{#2}{\usebox{\arx@rbox}}%
  \else\usebox{\arx@rbox}\fi}
\makeatother

\title{DominoTree: Conditional Tree-Structured Drafting with Domino for
Speculative Decoding\thanks{Code: \url{https://github.com/slin-zhq/Domino-Tree}}}

\author{
  Saw S. Lin (Zhiqi Zhang) \and Jyh-Shing Roger Jang
}
\date{}

\begin{document}
\maketitle

\begingroup
\renewcommand\thefootnote{}
\footnotetext{
Department of Computer Science and Information Engineering, National Taiwan University.
Correspondence to: Saw S. Lin (Zhiqi Zhang) <\texttt{r13922176@csie.ntu.edu.tw}>.\\
This work has been submitted to the IEEE for possible publication. Copyright may be transferred
without notice, after which this version may no longer be accessible.
}
\endgroup

\vspace{-2.5em}

\begin{abstract}
Speculative decoding speeds up large language model inference: a small drafter proposes several
tokens, and the target model verifies them in one parallel pass. Drafting a tree lets the target verify
many alternative continuations at once. DDTree builds such trees from the block-diffusion drafter DFlash,
but scores each position independently of the tokens earlier in its branch; Domino conditions on those
tokens but drafts only a single chain.
DominoTree combines the two: a training-free best-first tree scored by Domino's path-dependent
causal correction, restricted to top-M candidates. In a research harness at temperature zero, it
averages 5.33x speedup over autoregressive (AR) decoding on Qwen3-4B and 6.89x on Qwen3-8B,
peaking at 8.77x; across temperatures it is 20--26\% faster than Domino on Qwen3-4B and 27--36\% on Qwen3-8B,
and 11--20\% faster than the stronger of DDTree and its cost-aware variant CaDDTree. A controlled decomposition attributes
+10.1\% (Qwen3-4B) and +10.7\% (Qwen3-8B) accepted length to the correction, and +4.7\% and
+4.4\% to conditioning on the realized path. Served in SGLang on one RTX 5090, it reaches 4.11x and
4.83x AR at batch size 1, 1.17x and 1.09x the speed of Domino's chain, and up to 1.71x AR on
8K-token HELMET summarization, 14--21\% above the chain. The lead holds at 2 concurrent requests
but reverses from 8 onward, where the chain sustains more goodput despite DominoTree's 28\% higher acceptance. DominoTree
accepts more tokens than the chain in every serving setting.
On the hybrid Gated-DeltaNet Qwen3.6-27B, correct output requires tree-aware recurrent
verification; with it, DominoTree matches the chain at batch size 1 (4.58x versus 4.54x AR) and
reaches up to 2.11x AR at long context, 15\% above the chain.
\end{abstract}

\section{Introduction}
\label{sec:intro}

\providecommand{\PARstart}[2]{#1#2}
\PARstart{E}{very} speculative decoding method trades draft quality against draft cost:
end-to-end speedup over autoregressive (AR) decoding grows with the tokens accepted per round
($\tau$) and shrinks with the time spent drafting and verifying (\eqref{eq:speedup},
Section~\ref{sec:background-specdec}). Autoregressive drafters such as the EAGLE
line~\citep{li2024eagle,li2024eagle2,li2025eagle3} condition each draft token on the ones
before it, at the cost of one sequential forward pass per token. Block-diffusion drafters such
as DFlash~\citep{chen2026dflash} propose a whole block in one parallel pass, but each
position's logits are a \emph{marginal} that ignores the tokens realized elsewhere in the
block, which caps $\tau$ (Section~\ref{sec:background-dflash}).

Domino~\citep{huang2026domino} narrows this gap cheaply. It keeps DFlash's parallel backbone
and adds a lightweight sequential correction: a gated recurrent unit (GRU) tracks the tokens
sampled so far along the draft and adjusts each position's logits toward the true
conditional (Section~\ref{sec:background-domino}). The released implementation, however, uses
the correction to walk a single path per round and verifies one chain; it never branches.

\begin{figure*}[t]
\centering
\includegraphics[width=\textwidth]{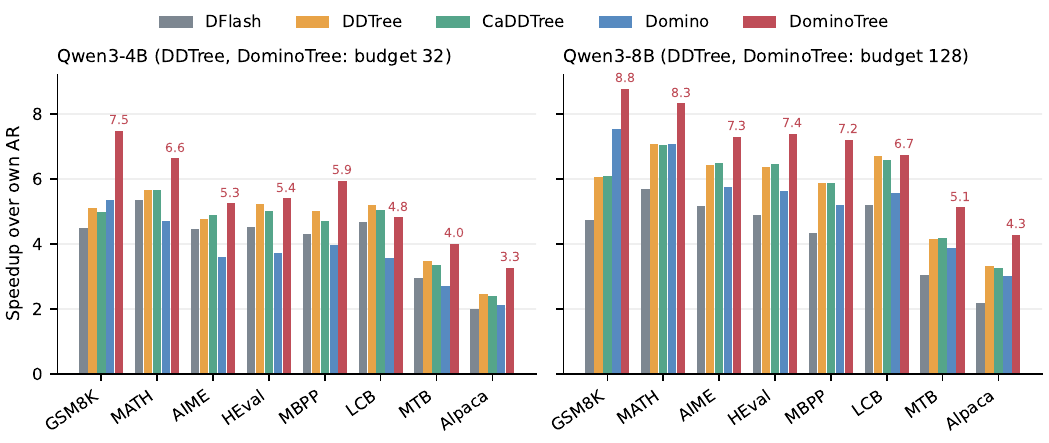}
\caption{Speedup over AR at $T{=}0$ across the eight datasets, Qwen3-4B (left, tree budget 32) and
Qwen3-8B (right, tree budget 128). Accepted lengths are in Table~\ref{tab:main}.}
\label{fig:speedup-preview}
\end{figure*}

Tree drafting, which verifies several candidate continuations in one target forward pass, is
the standard way to raise $\tau$ further, and DDTree~\citep{ringel2026ddtree} and its
cost-aware extension CaDDTree~\citep{zhang2026caddtree} build such trees on DFlash
(Sections~\ref{sec:background-ddtree}--\ref{sec:background-caddtree}). Both assume the draft
distribution \emph{factorizes}: a token's score at position $i$ depends on $i$ alone, not on
the tokens chosen before it in the tree. This holds for DFlash's marginals and fails for
Domino, which is \emph{partial-conditional}: its backbone states are path-independent, but its
GRU correction is not (Section~\ref{sec:background-domino}). Applied to Domino
directly, DDTree either falls back to Domino's single path or discards the correction.

We introduce DominoTree, a training-free method that keeps DDTree's best-first heap but scores
each node by re-running Domino's released GRU correction along that node's own root-to-node
path. On Qwen3-8B at $T{=}0$, DominoTree averages a $6.89\times$ speedup over AR and $10.43$
accepted tokens per round, and it has the highest mean accepted length of all evaluated
methods (DFlash, DDTree, CaDDTree, Domino) at every temperature tested (Table~\ref{tab:main}).

\paragraph{Contributions.}
\begin{itemize}
  \item \textbf{DominoTree} (Section~\ref{sec:method}). Restricting the per-node correction to
    the marginal top-$M$ candidates at each depth makes conditional scoring affordable. Across
    $T\in\{0.0,0.5,1.0\}$, DominoTree is $19.81$--$26.05\%$ faster than Domino's fastest
    (CUDA-graph) configuration on Qwen3-4B and $27.15$--$36.41\%$ on Qwen3-8B
    (Table~\ref{tab:pairwise}), and its paired gains over DDTree and CaDDTree are positive,
    with confidence intervals excluding zero, in every task family, model and temperature.
  \item \textbf{A host-synchronization-free tree builder} (Section~\ref{sec:method-frontier}).
    A depth-synchronous \emph{frontier} builder advances $W$ lanes per request in lockstep,
    records every scored candidate in a ledger and selects the global top-$B$ at the end, with
    no host synchronization. It provably selects the same nodes as the sequential heap (up to
    exact ties), so it changes only the cost of building the tree: $5.2$--$8.3\times$ less
    construction time (Section~\ref{sec:exp-builder}). It produces every headline and served
    result. %
  \item \textbf{A negative result for adaptive budgets}
    (Appendix~\ref{sec:exp-condadaptive}). Ported to this tree, CaDDTree's cost-aware budget
    rule runs to its cap, because the corrected drafter's path probabilities over-credit
    acceptance; we therefore use a fixed budget and report the calibration evidence.
  \item \textbf{A controlled decomposition} (Appendix~\ref{sec:exp-c3}). An intermediate arm that
    applies Domino's correction once per depth separates the value of the correction from the
    value of recomputing it per path: accepted length rises $6.92 \to 7.62 \to 7.98$ on
    Qwen3-4B ($+10.1\%$, then $+4.7\%$) and by $+10.7\%$ and $+4.4\%$ on Qwen3-8B, with both
    steps positive on every dataset.
  \item \textbf{Production serving and long context} (Section~\ref{sec:serving}). In SGLang on
    one RTX~5090, DominoTree has the highest accepted length in all $32$ batch-size-1 settings
    and beats the Domino chain's throughput in all $32$ ($27$ significant): $4.11\times$ and
    $4.83\times$ AR at $T{=}0$ on Qwen3-4B and Qwen3-8B, $1.17\times$ and $1.09\times$ the
    chain. Under load its acceptance lead stays near $28\%$, but goodput leads at $c{=}2$, is close to
    the chain at $c{=}4$ ($+0.2\%$ on Qwen3-4B, $-5.3\%$ on Qwen3-8B), and trails from $c{=}8$. On HELMET (How to Evaluate Long-Context Language Models
    Effectively and Thoroughly)~\citep{helmet2025} summarization at 8K--32K it accepts more
    than the chain in all $12$ settings and is significantly faster in $10$. With tree-aware
    recurrent verification it also serves the hybrid Qwen3.6-27B (Section~\ref{sec:serving-27b}).
\end{itemize}

The best-first selection rule follows DDTree, and tree-attention verification is reused
unchanged (Section~\ref{sec:method-verify}). Our contribution is the path-dependent node score,
the frontier builder that computes the same tree without per-node host synchronization, and
tree-aware recurrent verification for the hybrid target. %

\section{Background}
\label{sec:background}

\subsection{Speculative decoding}
\label{sec:background-specdec}

Speculative decoding~\citep{leviathan2023speculative,chen2023accelerating} accelerates
autoregressive large language model (LLM) inference by pairing a cheap \emph{draft} model with
the expensive \emph{target} model. Each round, the drafter proposes candidate tokens (a chain,
or a tree in tree-based methods); the target scores all of them in one parallel forward pass and
accepts the longest prefix, or root-to-leaf path, that matches its own choices. The output
distribution is the target's own, so the method is \emph{lossless up to floating-point
tie-breaking}. This caveat applies to any method that verifies several positions in one pass, chain
or tree: the pass splits the same bf16 dot products into partial sums differently than
token-by-token decoding does, and where two candidates score almost identically the rounding can
reverse which one is larger. This is a property of
finite-precision arithmetic, not an approximation introduced by the tree.

End-to-end speedup over AR decoding is
\begin{equation}
  \eta = \frac{\tau \cdot L_{\text{target}}}{T_{\text{draft}} + T_{\text{verify}}},
  \label{eq:speedup}
\end{equation}
where $\tau$ is the mean number of tokens accepted per round (the \emph{acceptance length}),
$L_{\text{target}}$ the cost of one target forward pass, and $T_{\text{draft}}$ and
$T_{\text{verify}}$ the per-round drafting and verification costs. Draft \emph{quality} raises
$\tau$ and draft \emph{cost} raises $T_{\text{draft}}$; a method that buys $\tau$ with a
proportionally larger $T_{\text{draft}}$ gains nothing. An autoregressive drafter conditions
every token on the previous ones but needs one sequential pass per drafted token; parallel
(block) drafters give that up.

\subsection{DFlash: a block-diffusion drafter and why its logits are marginals}
\label{sec:background-dflash}

DFlash~\citep{chen2026dflash} drafts a block of $B{-}1$ tokens in one non-causal forward pass.
Given the last verified token $x_t$, its backbone reads the masked block
$[x_t, \texttt{[MASK]}, \dots, \texttt{[MASK]}]$ together with features from the target's hidden
states, and produces one hidden state $H_i$ per draft position. The frozen target
language-modeling (LM) head turns each into base logits
\[
  L_i^{\text{base}} = \mathrm{LMHead}(H_i).
\]
Because every position sees the same masked block, $L_i^{\text{base}}$ estimates the
\emph{marginal} $p(x_{t+i}\mid x_{\le t})$, not the conditional
$p(x_{t+i}\mid x_{\le t}, x_{t+1},\dots,x_{t+i-1})$ the target uses. Errors from sampling
positions independently compound along the block, which is the structural source of the
quality gap between block-diffusion and autoregressive drafters.

\subsection{Domino: a partial-conditional drafter}
\label{sec:background-domino}

Domino~\citep{huang2026domino} recovers much of this gap with a lightweight sequential
correction on top of DFlash's unchanged backbone. A single-layer GRU accumulates a state from
the (frozen) target embeddings $E_j$ of the tokens actually sampled on the current path,
\[
  S_{i-1} = \mathrm{GRU}(E_{\le i-1}),
\]
and a low-rank multilayer perceptron (MLP) turns the backbone state and this causal state into
a logit-space correction,
\[
  \Delta L_i = W_2\,\sigma\!\big(W_1\,[H_i; S_{i-1}]\big),
\]
where $W_1$ projects to a rank-256 bottleneck, $W_2$ projects back to the full vocabulary and
$\sigma$ is SiLU. The draft logits are $L_i = L_i^{\text{base}} + \Delta L_i$, and the token
sampled from them is fed back into the GRU. Correcting in logit space keeps the expensive LM
head to one call per round. Both components are released weights, so everything we build on
them is training-free.

Two properties of Domino shape how a tree can be built on it. First, the released decoder is
chain-only: it samples one token per position and advances a single GRU state. Nothing in the
architecture prevents branching (each candidate could carry its own state), but the release does
not branch. Second, Domino conditions only partially. Its backbone state $H_i$ is computed once and is the
same whatever tokens are sampled; only the correction depends on the path, through $S_{i-1}$:
\begin{align*}
  q_{\text{Domino}}\big(x_{t+i} \mid x_{\le t},\, x_{t+1:t+i-1}\big)
    &= \mathrm{softmax}\Big(L_i^{\text{base}} \\
  &{}+ \Delta L_i\big(H_i,\, S_{i-1}(x_{t+1:t+i-1})\big)\Big).
\end{align*}
Domino therefore sits between the two regimes trees are usually built for: an autoregressive drafter
must re-run its backbone per branch, and a marginal drafter's branches cost only a top-$k$ read per
depth. Branching on Domino re-runs only the cheap GRU and correction per node, never the backbone.
This is what makes a conditional tree on Domino affordable, and it is the dependence that DDTree's
and CaDDTree's factorized formulation cannot represent.

\subsection{DDTree: a best-first heap over factorized marginals}
\label{sec:background-ddtree}

DDTree~\citep{ringel2026ddtree} builds a draft tree from a block-diffusion drafter's marginals.
Under a node budget $n$, it keeps a max-heap keyed by cumulative path log-probability:
starting from the root's children, it repeatedly pops the highest-scoring node, adds it to the
tree and pushes that node's children, scored by the marginal at the child's depth. The scores
are \emph{path-independent}: every node at a given depth is offered the same children with the
same scores. The tree is prefix-closed by construction and is verified in one target pass with
an ancestor-only attention mask, the tree-attention mechanism of DDTree and
SpecInfer~\citep{miao2024specinfer}, which we reuse unchanged (Section~\ref{sec:method-verify}).

\subsection{CaDDTree: cost-aware adaptive budget}
\label{sec:background-caddtree}

A larger tree never accepts fewer tokens in expectation, so acceptance alone always favors the
largest affordable budget. CaDDTree~\citep{zhang2026caddtree} instead maximizes expected
throughput. With $\Phi(n) = \sum_{s} \pi_s$ the prefix-probability mass of the first $n$ nodes
in best-first order, a fixed draft cost $C_d$ and an increasing, convex verification cost
$C_v(n)$,
\[
  \theta(n) = \frac{1 + \Phi(n)}{C_d + C_v(n)}
\]
is unimodal in $n$, so the tree can grow node by node and stop the first time $\theta$ stops
improving. This adapts the budget to each round without an offline search. The proof assumes
the factorized case; Appendix~\ref{sec:exp-condadaptive} applies the rule to a conditional tree
as a heuristic and reports the result.

\section{Methodology}
\label{sec:method}

DominoTree builds a draft tree on the released Domino checkpoint without retraining or changing
its weights. It has four parts: DDTree's best-first heap, scored with Domino's GRU-corrected
conditional log-probabilities instead of marginals (Section~\ref{sec:method-heap}); a candidate
restriction that makes the per-node correction affordable (Section~\ref{sec:method-topm}); a
frontier builder that computes the same tree as the sequential heap with no host synchronization
(Section~\ref{sec:method-frontier}); and tree-attention verification, borrowed unchanged
(Section~\ref{sec:method-verify}). The hybrid Qwen3.6-27B target also needs tree-aware recurrent
verification (Section~\ref{sec:serving-27b}). An adaptive-budget variant was tried and not
adopted (Appendix~\ref{sec:exp-condadaptive}).

\subsection{Conditional-scored best-first heap}
\label{sec:method-heap}

The builder is DDTree's algorithm unchanged: a max-heap keyed by cumulative path
log-probability, popped one node at a time in non-increasing score order, each popped node
pushing its children, until the node budget $n$ or the maximum depth $L$ is reached. Everything
that differs from DDTree on Domino lies in one function, the \emph{children function}, which the
heap calls to score a popped node's children:
\[
  \mathcal{C}(\text{state},\ d) \;\longrightarrow\;
  (\text{tokens},\ \text{log-probs},\ \text{child-states}).
\]
Given a node's path state (the GRU state after the tokens from the root to that node) and the
next depth, $\mathcal{C}$ returns the top-$k$ candidate tokens, their log-probabilities and, for
the conditional scorer, each child's own state. Both scorers plug into the same builder through
this interface alone, so Appendix~\ref{sec:exp-c3} can vary the score and nothing else.

\paragraph{Marginal children (DDTree's scorer).} This scorer ignores the path state and returns
the top-$k$ tokens of Domino's base logits at depth $d$:
\[
  \mathcal{C}_{\mathrm{marg}}(\cdot,\ d) = \text{top-}k\big(L_d^{\text{base}}\big).
\]
It is computed once per depth and shared by every node at that depth. We use it only as the
ablation control Marg@16 (Appendix~\ref{sec:exp-c3}); it is not a contribution.

\paragraph{Conditional children (our scorer).} This scorer re-runs Domino's released correction
head for the specific path that led to the node. Given the path's GRU state $S_{i-1}$, carried
in the heap with each entry, and the shared backbone state $H_i$,
\[
  \mathcal{C}_{\mathrm{cond}}(S_{i-1},\ i) =
    \text{top-}k\Big(L_i^{\text{base}} + \Delta L_i(H_i, S_{i-1})\Big),
\]
with $\Delta L_i = W_2\,\sigma(W_1[H_i; S_{i-1}])$ as in Section~\ref{sec:background-domino}.
The GRU is then advanced once per selected token, giving each child the state that will score
\emph{its} children. Because $H_i$ is shared across paths (Section~\ref{sec:background-domino}), a wider or deeper tree needs
no extra backbone computation; only the correction and the GRU, both cheap next to the backbone's
vocabulary projection, run per node.

\paragraph{Non-factorization of the conditional score.} $\mathcal{C}_{\mathrm{marg}}$ is
\emph{factorized}: every node at depth $d$ is offered the same menu. $\mathcal{C}_{\mathrm{cond}}$
is not, because $\Delta L_i$ depends on the path through $S_{i-1}$. This places DominoTree outside
the factorized case that DDTree's correctness argument and CaDDTree's unimodality proof assume.
Our claim is not a new guarantee but that Domino's partial-conditional structure makes such a
non-factorized score cheap enough to drive DDTree's heap unchanged.

\subsection{Candidate restriction and the best-first construction}
\label{sec:method-topm}

The conditional scorer's dominant cost is $W_2$, which maps the 256-dimensional bottleneck to
Qwen3's $151{,}936$-token vocabulary, once per node and one node at a time, since best-first
expansion pops sequentially. Left unrestricted, this projection dominates build time and
consumes the acceptance gain (Table~\ref{tab:topm-ablation}). We therefore correct only the
marginal top-$M$ tokens at each depth, $M \ll |V|$: once per depth, gather the $M$ rows of $W_2$
and the base log-probabilities of the top-$M$ marginal tokens; per node, correct only that slice
($256\times M$ instead of $256\times|V|$) and take its top-$k$. The expansion order and the
tree's depth are unaffected. Algorithm~\ref{alg:dominotree} gives the full builder.

\begin{myalg}
\algocaption{Best-first conditional draft-tree construction}
\label{alg:dominotree}
\small
\textbf{Require:} Backbone hidden states $\{H_i\}_{i=1}^{L}$ and base
logits $\{L_i^{\text{base}}\}_{i=1}^{L}$ from one Domino forward pass; root
GRU state $S_0$; node budget $n$; candidate width $M$.\\
\textbf{Heap entries.} Each entry of the max-heap $\mathcal{H}$ is a triple
(token sequence, cumulative corrected log-score $g$, GRU state); $u \Vert v$
denotes the sequence $u$ extended by the single token $v$.
\begin{enumerate}
\item \textbf{Precompute}, once per depth $d$: the top-$M$ marginal token
  indices under $L_d^{\text{base}}$ and the corresponding $M$ rows of $W_2$.
\item \textbf{function} \textsc{Children}($S_{d-1}$, $d$) --- \emph{applies
  the correction; called once per node expansion, for the root (Step~3) and
  every interior pop (Step~4) alike}:
  \begin{enumerate}
  \item $\Delta L_d \gets W_2\,\sigma\big(W_1[H_d; S_{d-1}]\big)$,
    restricted to the depth-$d$ top-$M$ slice.
  \item \textbf{return} $k$ triples, one per selected token: the token, its
    \emph{corrected} log-probability $\ell_d(\cdot)$ under
    $L_d^{\text{base}} + \Delta L_d$ restricted to that slice, and the GRU state
    obtained by advancing $S_{d-1}$ by that token.
  \end{enumerate}
\item Initialize $\mathcal{H}$ with the root's children: for every triple
  $(v, \ell_1(v), S_1)$ returned by $\textsc{Children}(S_0, 1)$, insert the
  entry $\big((v),\, \ell_1(v),\, S_1\big)$. Set $T \gets \varnothing$.
\item \textbf{while} $|T| < n$ and $\mathcal{H} \neq \varnothing$:
  \begin{enumerate}
  \item \textbf{Pop from} $\mathcal{H}$ the entry with the largest cumulative
    corrected log-score; write it $\big(u, g(u), S_d\big)$ with
    $u=(u_1,\dots,u_d)$. Add $u$ to $T$.
  \item \textbf{if} $d < L$: call $\textsc{Children}(S_d, d{+}1)$ ---
    recomputing $\Delta L_{d+1}$ for this specific path --- and, for every
    returned triple $(v, \ell_{d+1}(v), S_{d+1})$, \textbf{push onto}
    $\mathcal{H}$ the entry
    $\big(u \Vert v,\ g(u)+\ell_{d+1}(v),\ S_{d+1}\big)$.
  \end{enumerate}
\item \textbf{return} draft tree $T$.
\end{enumerate}
\end{myalg}

\subsection{Host-synchronization-free frontier construction}
\label{sec:method-frontier}

Candidate restriction removes most of the correction's arithmetic, but
Algorithm~\ref{alg:dominotree}'s pop loop is sequential: each pop reads its node back to the host
before the next node can be scored, so every pop pays several small kernel launches and a
device-to-host synchronization. On a GPU this per-pop overhead, not the arithmetic, sets the
floor on build time. Scoring a whole depth level in one launch does not fix it on its own: best-first
grows the tree \emph{deep} along the best paths, and keeping whole levels would spend the budget
on shallow breadth.

The \emph{frontier builder} keeps one batched launch per depth without paying for breadth. It
advances $W = B$ lanes per request in lockstep (with $B$ the node budget), scores every lane's $k$
children with the conditional correction in one batched operation that also spans every in-flight
request, and continues with the top $W$. Every scored candidate, kept or not, is written to a
fixed-shape ledger with its cumulative log-probability and a parent pointer. After the last depth
the tree is selected once, as the ledger's global top $B$. This is the heap's tree: a child never
scores above its parent, so the $B$ best nodes form an ancestor-closed set, which is what
best-first expansion pops, and with $W = B$ no such node is dropped before its children are scored
(Appendix~\ref{app:gpu-builder} gives the argument, the case $W < B$, and the effect of bf16 near-ties). The builder therefore
changes only the cost of construction. It has no host synchronization, is captured as one CUDA graph
per batch size, produces every headline and serving DominoTree result, and is compared with the
Python heap in Section~\ref{sec:exp-builder}.

Two further changes reduce the builder's cost and, apart from rare near-exact ties, do not change
the selected tree. Each depth's scoring, selection and GRU update run as a few fused Triton kernels
instead of many small PyTorch kernels, because at these tree sizes the number of launches, not the
arithmetic, sets build time. A GRU input table precomputes once, for every vocabulary
token, its embedding multiplied by the GRU's fixed input weights $W_{ih}$, so each depth gathers a
row instead of doing an embedding lookup and a matrix multiply. It is enabled in every
frontier-builder result we report; it needs an unsplit embedding, so it cannot be used under tensor
parallelism, which none of those results uses. Both reorder floating-point reductions, so neither is
bit-identical to the unfused code, but pooled accepted length stays within half a percent of it
(Appendix~\ref{app:gpu-builder}). Table~\ref{tab:builder-stagetime} reports the build times.

\subsection{Tree-attention verification (borrowed, unmodified)}
\label{sec:method-verify}

The tree is verified with the tree attention of DDTree and SpecInfer~\citep{miao2024specinfer},
unchanged: an ancestor-only causal mask lets one target forward pass score every position, and the
longest root-to-leaf path whose tokens are all accepted is committed. This runs alongside, and does
not modify, Domino's own chain verification; its nearly flat cost as the tree grows is inherited,
not engineered here. For the hybrid Qwen3.6-27B target the attention mask alone is not enough, and
Section~\ref{sec:serving-27b} adds tree-aware recurrent verification.

\section{Experiments}
\label{sec:experiments}

\subsection{Setup}
\label{sec:exp-setup}

\paragraph{Models and hardware.} The targets are Qwen3-4B and Qwen3-8B~\citep{yang2025qwen3}
with the released Domino drafters \texttt{Qwen3-4B-Domino-b16} and \texttt{Qwen3-8B-Domino-b16}
(block size 16, prefix length 1). DominoTree runs in our single-stream, batch-size-1 research
harness on the HuggingFace (HF) Transformers backend~\citep{wolf2020transformers}; DFlash, DDTree
and CaDDTree run in the official CaDDTree implementation (the reference harness) on the native
\texttt{-DFlash-b16} drafters; the released Domino decoder runs in its own published benchmark.
Qwen3-4B runs on RTX~5080 GPUs (16\,GB), one per harness; every Qwen3-8B method, at every
temperature, runs in one session on a single H100 (80\,GB). Because absolute throughput differs
across platforms, we compare speedups over the AR baseline of the same platform.

\paragraph{Datasets.} We use the eight datasets of Domino's evaluation: GSM8K~\citep{cobbe2021gsm8k},
MATH-500~\citep{lightman2024verify} (from MATH~\citep{hendrycks2021math}) and
AIME25~\citep{balunovic2025matharena} for math; HumanEval~\citep{chen2021humaneval},
MBPP~\citep{austin2021mbpp} (sanitized) and LiveCodeBench (LCB)~\citep{jain2025livecodebench} for
code; and MT-Bench~\citep{zheng2023mtbench} and Alpaca~\citep{taori2023alpaca} for chat. MT-Bench
runs as a two-turn conversation, each method feeding its own first answer into the second turn.

\paragraph{Generation length and sample size.} We generate up to $2048$ tokens per prompt, as in
Domino's Table~1 and the reference protocol, over $50$ prompts per dataset ($30$ for AIME25).
Every harness except official Domino's runs a warm-up generation before timing, so all 430
prompt/turn rows per model, temperature and method are measured warm. Official Domino's benchmark
has no warm-up, so we drop its first prompt on each dataset (421 rows remain) and leave out
DominoTree's matching prompt in paired comparisons against it.

\paragraph{Temperatures.} We evaluate $T\in\{0.0,0.5,1.0\}$; $T{=}0$ is greedy. For $T{>}0$,
DominoTree samples the target while keeping the draft deterministic, as the reference DDTree and
CaDDTree harness does, so the tree comparison is like for like; the released Domino decoder
follows its own published decoding. Draft-side sampling is ablated in
Table~\ref{tab:draftsample}. Every method is lossless at every temperature.

\paragraph{Baselines: a three-harness protocol.} (1)~The reference harness (CaDDTree commit
\texttt{a88f3f3}) supplies AR, DFlash~\citep{chen2026dflash},
DDTree (Section~\ref{sec:background-ddtree}) and CaDDTree
(Section~\ref{sec:background-caddtree}), with DDTree at node budgets 32 (Qwen3-4B) and 128
(Qwen3-8B). (2)~The released decoder runs Domino (single chain), taking the faster of
eager and CUDA-graph execution per setting, in the same session as our harness. (3)~Our harness
runs DominoTree with the fused frontier builder at budgets 32 and 128, candidate width
$M{=}64$ and $k{=}8$ children per node. Each budget is fixed across datasets and temperatures,
chosen by a $T{=}0$ sweep: 32 is the fastest of 16, 32 and 64 on Qwen3-4B
(Table~\ref{tab:budget-ablation}), and 128 the fastest of 16 to 256 on Qwen3-8B (MATH-500,
HumanEval and MT-Bench, 20 prompts each). Speedups are normalized to each harness's own AR run.
The reference harness times AR in every Qwen3-4B setting but only at $T{=}0$ for Qwen3-8B, and
DominoTree and official Domino share one $T{=}0$ AR run per model. AR runs one forward pass per
token at any temperature; where it was timed at every temperature, its throughput stayed within
$1.2\%$ of the $T{=}0$ value.

\paragraph{The serving harness (SGLang).} Section~\ref{sec:serving} runs DominoTree inside a
production engine. We package it as an out-of-tree speculative-decoding plugin for stock
SGLang~\citep{zheng2024sglang} that registers a \texttt{DOMINOTREE} algorithm, reuses SGLang's
EAGLE tree verification and vendors Domino's correction head, byte-identical to the release apart
from import paths and one declared tensor-parallel safety patch that has no effect on a single
GPU.\footnote{The patch makes a per-rank memory check group-wide, so tensor-parallel ranks cannot
disagree and deadlock. A verifier reverts every declared patch before hashing, so an undeclared
change still fails.} The served builder is the frontier builder
(Section~\ref{sec:method-frontier}), since the sequential heap does not survive continuous
batching. Five methods are compared: AR, DFlash (block size 16),
EAGLE-3~\citep{li2025eagle3}, Domino (the chain) and DominoTree; Domino
and DominoTree share the Domino drafter, and EAGLE-3 uses its released head. For fairness every
method runs its own CUDA graph under identical serving flags (SGLang's stock graph configuration,
\texttt{-{}-mem-fraction-static 0.8}). Serving runs on one RTX~5090 (32\,GB) without tensor
parallelism. DominoTree uses budget $32$ at batch size 1 and under concurrency, and $16$ (4B) and
$32$ (8B) for long context. Throughput is output tokens/s over the timed window, speedup is over
served AR, and $\tau$ is SGLang's mean accepted length.

\paragraph{What the served Domino baseline is.} The served Domino row is the official
drafter and correction head driven through \emph{our} plugin, not the released Domino fork, which
pins an older SGLang and cannot run beside the other methods on one engine. This makes
tree-versus-chain the tightest control in the paper (drafter, head, engine, scheduler, key-value
(KV) cache and verification are fixed; only the draft structure changes), but it is not Domino's
own serving system, and any integration overhead is charged to the baseline.
Table~\ref{tab:main}'s Domino row, in contrast, is the released decoder in its own benchmark.

\subsection{Main results}
\label{sec:exp-table1}

\begin{table*}[t]
\centering
\scriptsize
\setlength{\tabcolsep}{2.6pt}
\caption{Speedup over autoregressive (AR) decoding and mean accepted length $\tau$ on the eight
datasets, Qwen3-4B (one RTX~5080) and Qwen3-8B (one H100). Speedup is mean per-row tokens per
second (TPS) over AR's, normalized within each harness (Section~\ref{sec:exp-setup}).
DominoTree and DDTree use node budgets 32 (4B) and 128 (8B); CaDDTree chooses its budget each
round. Overall is the unweighted mean over datasets. Bold marks the highest speedup and,
separately, the highest $\tau$ per dataset.} %
\label{tab:main}
\resizebox{\textwidth}{!}{%
\begin{tabular}{ll*{9}{cc}}
\toprule
Model & Method & \multicolumn{6}{c}{Math} & \multicolumn{6}{c}{Code} & \multicolumn{4}{c}{Chat} & \multicolumn{2}{c}{Overall} \\
\cmidrule(lr){3-8}\cmidrule(lr){9-14}\cmidrule(lr){15-18}\cmidrule(lr){19-20}
 & & \multicolumn{2}{c}{GSM8K} & \multicolumn{2}{c}{MATH-500} & \multicolumn{2}{c}{AIME25} & \multicolumn{2}{c}{HumanEval} & \multicolumn{2}{c}{MBPP} & \multicolumn{2}{c}{LCB} & \multicolumn{2}{c}{MT-Bench} & \multicolumn{2}{c}{Alpaca} & \multicolumn{2}{c}{Avg.} \\
\midrule
\multicolumn{2}{c}{Temperature = 0} & Speedup & $\tau$ & Speedup & $\tau$ & Speedup & $\tau$ & Speedup & $\tau$ & Speedup & $\tau$ & Speedup & $\tau$ & Speedup & $\tau$ & Speedup & $\tau$ & Speedup & $\tau$ \\
\midrule
\multirow{5}{*}{Qwen3-4B}
& DFlash & 4.48 & 6.59 & 5.34 & 8.11 & 4.47 & 7.37 & 4.53 & 6.77 & 4.30 & 6.25 & 4.69 & 7.42 & 2.96 & 4.56 & 2.00 & 2.96 & 4.10 & 6.25 \\
& DDTree (32) & 5.10 & 8.01 & 5.66 & 9.24 & 4.77 & 8.51 & 5.24 & 8.45 & 5.00 & 7.76 & \textbf{5.19} & \textbf{8.90} & 3.48 & 5.73 & 2.47 & 3.93 & 4.61 & 7.57 \\
& CaDDTree & 4.97 & 7.47 & 5.65 & 8.70 & 4.88 & 8.22 & 5.02 & 7.70 & 4.70 & 7.03 & 5.04 & 8.16 & 3.36 & 5.27 & 2.41 & 3.66 & 4.50 & 7.03 \\
& Domino & 6.62 & 10.19 & 5.63 & 9.05 & 4.22 & 7.47 & 4.42 & 7.05 & 4.77 & 7.29 & 4.18 & 7.05 & 3.22 & 5.21 & 2.54 & 3.88 & 4.45 & 7.15 \\
& \textbf{DominoTree (32)} & \textbf{7.39} & \textbf{12.00} & \textbf{6.57} & \textbf{11.10} & \textbf{5.24} & \textbf{9.77} & \textbf{5.42} & \textbf{8.96} & \textbf{5.96} & \textbf{9.46} & 4.82 & 8.61 & \textbf{4.00} & \textbf{6.75} & \textbf{3.27} & \textbf{5.22} & \textbf{5.33} & \textbf{8.98} \\
\midrule
\multirow{5}{*}{Qwen3-8B}
& DFlash & 4.73 & 6.60 & 5.69 & 7.91 & 5.17 & 7.17 & 4.90 & 6.79 & 4.34 & 6.07 & 5.20 & 7.24 & 3.03 & 4.22 & 2.18 & 3.13 & 4.40 & 6.14 \\
& DDTree (128) & 6.07 & 8.69 & 7.08 & 10.13 & 6.44 & 9.29 & 6.37 & 9.05 & 5.88 & 8.43 & 6.72 & 9.59 & 4.15 & 5.95 & 3.34 & 4.87 & 5.75 & 8.25 \\
& CaDDTree & 6.11 & 8.94 & 7.05 & 10.30 & 6.48 & 9.51 & 6.45 & 9.39 & 5.89 & 8.61 & 6.58 & 9.69 & 4.19 & 6.12 & 3.27 & 4.82 & 5.76 & 8.42 \\
& Domino & 7.54 & 10.06 & 7.09 & 9.46 & 5.75 & 7.71 & 5.62 & 7.55 & 5.20 & 7.06 & 5.56 & 7.46 & 3.86 & 5.20 & 3.02 & 4.10 & 5.46 & 7.32 \\
& \textbf{DominoTree (128)} & \textbf{8.77} & \textbf{13.28} & \textbf{8.33} & \textbf{12.60} & \textbf{7.29} & \textbf{11.19} & \textbf{7.40} & \textbf{11.14} & \textbf{7.20} & \textbf{10.88} & \textbf{6.74} & \textbf{10.13} & \textbf{5.13} & \textbf{7.76} & \textbf{4.29} & \textbf{6.46} & \textbf{6.89} & \textbf{10.43} \\
\midrule
\multicolumn{2}{c}{Temperature = 0.5} & Speedup & $\tau$ & Speedup & $\tau$ & Speedup & $\tau$ & Speedup & $\tau$ & Speedup & $\tau$ & Speedup & $\tau$ & Speedup & $\tau$ & Speedup & $\tau$ & Speedup & $\tau$ \\
\midrule
\multirow{5}{*}{Qwen3-4B}
& DFlash & 4.39 & 6.38 & 5.03 & 7.58 & 3.94 & 6.53 & 4.40 & 6.57 & 4.12 & 5.99 & 4.61 & 7.28 & 2.88 & 4.39 & 1.95 & 2.90 & 3.92 & 5.95 \\
& DDTree (32) & 4.98 & 7.84 & 5.58 & 9.08 & 4.38 & 7.77 & 5.00 & 8.04 & 5.01 & 7.76 & 4.87 & \textbf{8.26} & 3.46 & 5.68 & 2.44 & 3.89 & 4.46 & 7.29 \\
& CaDDTree & 4.80 & 7.21 & 5.57 & 8.59 & 4.32 & 7.31 & 4.95 & 7.57 & 4.62 & 6.92 & \textbf{4.91} & 7.94 & 3.38 & 5.29 & 2.38 & 3.63 & 4.37 & 6.81 \\
& Domino & 6.03 & 9.39 & 5.30 & 8.57 & 3.47 & 6.25 & 4.32 & 6.95 & 4.49 & 6.93 & 3.92 & 6.62 & 3.04 & 4.96 & 2.44 & 3.76 & 4.13 & 6.68 \\
& \textbf{DominoTree (32)} & \textbf{6.88} & \textbf{11.29} & \textbf{6.11} & \textbf{10.38} & \textbf{4.46} & \textbf{8.38} & \textbf{5.30} & \textbf{8.85} & \textbf{5.58} & \textbf{8.93} & 4.64 & 8.20 & \textbf{3.88} & \textbf{6.61} & \textbf{3.18} & \textbf{5.10} & \textbf{5.00} & \textbf{8.47} \\
\midrule
\multirow{5}{*}{Qwen3-8B}
& DFlash & 4.52 & 6.37 & 5.44 & 7.60 & 4.51 & 6.36 & 4.47 & 6.32 & 4.20 & 5.97 & 5.07 & 7.14 & 2.96 & 4.13 & 2.10 & 3.03 & 4.16 & 5.87 \\
& DDTree (128) & 5.88 & 8.55 & 6.76 & 9.76 & 5.78 & 8.53 & 6.11 & 8.87 & 5.65 & 8.28 & \textbf{6.54} & 9.49 & 4.04 & 5.87 & 3.14 & 4.64 & 5.49 & 8.00 \\
& CaDDTree & 5.83 & 8.72 & 6.77 & 9.99 & 5.91 & 8.90 & 6.05 & 9.05 & 5.66 & 8.50 & 6.51 & \textbf{9.77} & 4.02 & 5.98 & 3.15 & 4.74 & 5.49 & 8.21 \\
& Domino & 6.83 & 9.23 & 6.23 & 8.44 & 4.69 & 6.41 & 5.15 & 7.05 & 4.86 & 6.74 & 5.43 & 7.27 & 3.56 & 4.86 & 2.73 & 3.70 & 4.93 & 6.71 \\
& \textbf{DominoTree (128)} & \textbf{8.47} & \textbf{13.03} & \textbf{7.94} & \textbf{12.01} & \textbf{6.53} & \textbf{10.08} & \textbf{6.83} & \textbf{10.48} & \textbf{6.83} & \textbf{10.45} & 6.45 & \textbf{9.77} & \textbf{4.91} & \textbf{7.50} & \textbf{4.06} & \textbf{6.23} & \textbf{6.50} & \textbf{9.94} \\
\midrule
\multicolumn{2}{c}{Temperature = 1} & Speedup & $\tau$ & Speedup & $\tau$ & Speedup & $\tau$ & Speedup & $\tau$ & Speedup & $\tau$ & Speedup & $\tau$ & Speedup & $\tau$ & Speedup & $\tau$ & Speedup & $\tau$ \\
\midrule
\multirow{5}{*}{Qwen3-4B}
& DFlash & 4.11 & 5.99 & 4.38 & 6.60 & 2.94 & 4.93 & 4.15 & 6.20 & 4.00 & 5.82 & 4.16 & 6.55 & 2.70 & 4.15 & 1.93 & 2.87 & 3.55 & 5.39 \\
& DDTree (32) & 4.81 & 7.55 & 5.13 & 8.31 & \textbf{3.58} & 6.41 & 4.80 & 7.75 & 4.79 & 7.45 & 4.77 & \textbf{8.09} & 3.21 & 5.26 & 2.39 & 3.77 & 4.18 & 6.82 \\
& CaDDTree & 4.55 & 6.83 & 4.90 & 7.58 & 3.48 & 5.89 & 4.59 & 7.05 & 4.49 & 6.70 & \textbf{4.84} & 7.75 & 3.15 & 4.91 & 2.36 & 3.59 & 4.04 & 6.29 \\
& Domino & 5.17 & 8.05 & 4.60 & 7.47 & 2.63 & 4.70 & 3.88 & 6.24 & 4.22 & 6.49 & 3.79 & 6.34 & 2.76 & 4.48 & 2.19 & 3.37 & 3.65 & 5.89 \\
& \textbf{DominoTree (32)} & \textbf{6.48} & \textbf{10.56} & \textbf{5.33} & \textbf{9.03} & 3.49 & \textbf{6.59} & \textbf{4.93} & \textbf{8.21} & \textbf{5.43} & \textbf{8.65} & 4.51 & 7.92 & \textbf{3.64} & \textbf{6.19} & \textbf{3.06} & \textbf{4.88} & \textbf{4.61} & \textbf{7.76} \\
\midrule
\multirow{5}{*}{Qwen3-8B}
& DFlash & 4.36 & 6.05 & 4.67 & 6.45 & 3.47 & 4.89 & 4.03 & 5.68 & 3.85 & 5.40 & 4.59 & 6.53 & 2.63 & 3.75 & 2.02 & 2.90 & 3.70 & 5.21 \\
& DDTree (128) & 5.58 & 8.01 & 6.20 & 8.84 & 4.85 & 7.11 & 5.62 & 8.15 & 5.47 & 7.91 & 6.39 & 9.33 & 3.82 & 5.62 & 3.00 & 4.40 & 5.12 & 7.42 \\
& CaDDTree & 5.70 & 8.38 & 6.29 & 9.19 & \textbf{4.92} & \textbf{7.34} & 5.55 & 8.25 & 5.46 & 8.04 & 6.16 & 9.28 & 3.87 & 5.82 & 3.05 & 4.55 & 5.12 & 7.60 \\
& Domino & 6.04 & 8.10 & 5.15 & 6.91 & 3.37 & 4.56 & 4.42 & 5.96 & 4.35 & 5.95 & 5.16 & 6.93 & 3.00 & 4.10 & 2.52 & 3.41 & 4.25 & 5.74 \\
& \textbf{DominoTree (128)} & \textbf{7.53} & \textbf{11.39} & \textbf{6.68} & \textbf{10.17} & 4.69 & 7.23 & \textbf{6.09} & \textbf{9.21} & \textbf{6.33} & \textbf{9.57} & \textbf{6.48} & \textbf{9.74} & \textbf{4.35} & \textbf{6.68} & \textbf{3.88} & \textbf{5.85} & \textbf{5.75} & \textbf{8.73} \\
\bottomrule
\end{tabular}
}
\end{table*}

DominoTree has the highest overall accepted length and speedup at every temperature on both
models (Table~\ref{tab:main}), and every task-family gain over Domino, DDTree and CaDDTree is
positive with a confidence interval excluding zero (Table~\ref{tab:pairwise}). The gain is largest
where drafting is hardest. DominoTree is faster than Domino's chain in all 48 dataset--temperature
settings, by the widest margin on chat, where acceptance is lowest, and at higher temperature,
which lowers acceptance for every method: both make a chain lose the rest of its round at the first
mismatch more often, which is the case a tree hedges. Against DDTree and CaDDTree the margin is
widest on math and chat; the few settings where one of them leads are on LiveCodeBench and on
AIME25 at $T{=}1$.

\begin{table*}[!t]
\centering
\scriptsize
\caption{Throughput change $\Delta\%$ of DominoTree over each baseline, with 95\% confidence
interval (CI) from a paired bootstrap stratified by dataset and clustered by conversation
(10{,}000 draws); budgets as in Table~\ref{tab:main}. \emph{vs.\ Domino} compares pooled mean
per-row TPS; \emph{vs.\ DDTree} and \emph{vs.\ CaDDTree} compare speedups over each method's own
AR. Rollups pool rows. Comparisons with official Domino leave out each dataset's first prompt
(Section~\ref{sec:exp-setup}). Bold marks CIs wholly above zero.}

\label{tab:pairwise}
\begin{tabular}{l*{3}{cc}}
\toprule
Category & \multicolumn{2}{c}{vs.\ Domino} & \multicolumn{2}{c}{vs.\ DDTree} & \multicolumn{2}{c}{vs.\ CaDDTree} \\
 & $\Delta\%$ & 95\% CI & $\Delta\%$ & 95\% CI & $\Delta\%$ & 95\% CI \\
\midrule
\multicolumn{7}{c}{\textbf{Qwen3-4B}} \\
\midrule
\multicolumn{7}{l}{\textit{Temperature = 0}} \\
Math & \textbf{15.32} & $\mathbf{[13.24,\ 17.47]}$ & \textbf{25.57} & $\mathbf{[23.00,\ 28.28]}$ & \textbf{24.88} & $\mathbf{[20.44,\ 29.12]}$ \\
Code & \textbf{21.23} & $\mathbf{[18.87,\ 23.76]}$ & \textbf{4.95} & $\mathbf{[2.56,\ 7.51]}$ & \textbf{9.76} & $\mathbf{[6.99,\ 12.64]}$ \\
Chat & \textbf{25.05} & $\mathbf{[21.25,\ 28.98]}$ & \textbf{19.33} & $\mathbf{[15.68,\ 23.18]}$ & \textbf{23.44} & $\mathbf{[19.48,\ 27.60]}$ \\
\textbf{Overall} & \textbf{19.81} & $\mathbf{[18.34,\ 21.32]}$ & \textbf{15.77} & $\mathbf{[14.12,\ 17.45]}$ & \textbf{18.59} & $\mathbf{[16.42,\ 20.70]}$ \\
\midrule
\multicolumn{7}{l}{\textit{Temperature = 0.5}} \\
Math & \textbf{16.24} & $\mathbf{[13.02,\ 19.43]}$ & \textbf{18.82} & $\mathbf{[15.85,\ 21.85]}$ & \textbf{20.82} & $\mathbf{[17.70,\ 24.06]}$ \\
Code & \textbf{22.17} & $\mathbf{[19.09,\ 25.46]}$ & \textbf{4.35} & $\mathbf{[1.71,\ 7.08]}$ & \textbf{7.15} & $\mathbf{[4.51,\ 9.74]}$ \\
Chat & \textbf{27.89} & $\mathbf{[23.22,\ 32.92]}$ & \textbf{16.82} & $\mathbf{[12.51,\ 21.52]}$ & \textbf{19.49} & $\mathbf{[15.05,\ 24.01]}$ \\
\textbf{Overall} & \textbf{21.29} & $\mathbf{[19.30,\ 23.31]}$ & \textbf{12.57} & $\mathbf{[10.70,\ 14.44]}$ & \textbf{15.08} & $\mathbf{[13.18,\ 16.94]}$ \\
\midrule
\multicolumn{7}{l}{\textit{Temperature = 1}} \\
Math & \textbf{21.84} & $\mathbf{[17.97,\ 25.75]}$ & \textbf{15.07} & $\mathbf{[11.81,\ 18.49]}$ & \textbf{20.48} & $\mathbf{[16.73,\ 24.33]}$ \\
Code & \textbf{25.52} & $\mathbf{[22.26,\ 28.84]}$ & \textbf{3.47} & $\mathbf{[0.90,\ 6.02]}$ & \textbf{6.78} & $\mathbf{[3.96,\ 9.54]}$ \\
Chat & \textbf{33.04} & $\mathbf{[27.77,\ 38.86]}$ & \textbf{17.38} & $\mathbf{[13.22,\ 21.70]}$ & \textbf{19.40} & $\mathbf{[15.09,\ 23.85]}$ \\
\textbf{Overall} & \textbf{26.05} & $\mathbf{[23.78,\ 28.38]}$ & \textbf{10.92} & $\mathbf{[9.11,\ 12.85]}$ & \textbf{14.62} & $\mathbf{[12.60,\ 16.71]}$ \\
\midrule
\multicolumn{7}{c}{\textbf{Qwen3-8B}} \\
\midrule
\multicolumn{7}{l}{\textit{Temperature = 0}} \\
Math & \textbf{18.69} & $\mathbf{[16.61,\ 20.81]}$ & \textbf{26.31} & $\mathbf{[24.24,\ 28.42]}$ & \textbf{25.96} & $\mathbf{[23.79,\ 28.15]}$ \\
Code & \textbf{30.77} & $\mathbf{[27.49,\ 34.15]}$ & \textbf{12.49} & $\mathbf{[10.11,\ 15.01]}$ & \textbf{12.70} & $\mathbf{[10.26,\ 15.26]}$ \\
Chat & \textbf{35.76} & $\mathbf{[30.55,\ 41.51]}$ & \textbf{24.82} & $\mathbf{[21.40,\ 28.33]}$ & \textbf{24.63} & $\mathbf{[21.45,\ 27.92]}$ \\
\textbf{Overall} & \textbf{27.15} & $\mathbf{[25.27,\ 29.03]}$ & \textbf{20.44} & $\mathbf{[18.93,\ 21.98]}$ & \textbf{20.37} & $\mathbf{[18.84,\ 21.85]}$ \\
\midrule
\multicolumn{7}{l}{\textit{Temperature = 0.5}} \\
Math & \textbf{27.99} & $\mathbf{[25.16,\ 30.92]}$ & \textbf{26.23} & $\mathbf{[23.70,\ 28.85]}$ & \textbf{25.98} & $\mathbf{[23.59,\ 28.47]}$ \\
Code & \textbf{31.02} & $\mathbf{[27.47,\ 34.50]}$ & \textbf{9.91} & $\mathbf{[7.68,\ 12.19]}$ & \textbf{10.28} & $\mathbf{[7.88,\ 12.58]}$ \\
Chat & \textbf{41.53} & $\mathbf{[35.44,\ 48.22]}$ & \textbf{23.57} & $\mathbf{[19.82,\ 27.27]}$ & \textbf{23.93} & $\mathbf{[20.61,\ 27.29]}$ \\
\textbf{Overall} & \textbf{32.39} & $\mathbf{[30.21,\ 34.65]}$ & \textbf{19.03} & $\mathbf{[17.46,\ 20.61]}$ & \textbf{19.19} & $\mathbf{[17.65,\ 20.70]}$ \\
\midrule
\multicolumn{7}{l}{\textit{Temperature = 1}} \\
Math & \textbf{28.65} & $\mathbf{[24.74,\ 32.78]}$ & \textbf{15.84} & $\mathbf{[12.64,\ 19.15]}$ & \textbf{13.91} & $\mathbf{[11.16,\ 16.67]}$ \\
Code & \textbf{36.64} & $\mathbf{[32.55,\ 40.94]}$ & \textbf{8.07} & $\mathbf{[5.48,\ 10.79]}$ & \textbf{10.04} & $\mathbf{[7.17,\ 12.97]}$ \\
Chat & \textbf{47.94} & $\mathbf{[41.71,\ 54.40]}$ & \textbf{18.24} & $\mathbf{[13.34,\ 22.90]}$ & \textbf{16.61} & $\mathbf{[11.15,\ 21.66]}$ \\
\textbf{Overall} & \textbf{36.41} & $\mathbf{[33.77,\ 39.00]}$ & \textbf{13.27} & $\mathbf{[11.24,\ 15.26]}$ & \textbf{13.04} & $\mathbf{[10.98,\ 15.07]}$ \\
\bottomrule
\end{tabular}
\end{table*}

\subsection{Builder cost: Python reference vs.\ frontier builder}
\label{sec:exp-builder}
\label{sec:exp-table2}

On Qwen3-4B, verification costs five to seven times as much as drafting per round, and tree
construction (build time) is the one drafter-side cost that grows with the tree. Since the
frontier builder and the Python reference (Algorithm~\ref{alg:dominotree}) select the same nodes,
comparing them isolates construction cost. The Python builder's cost grows with the number of
nodes, the frontier builder's only with depth, so the frontier builder is five to eight times
faster, increasingly so at larger budgets (Table~\ref{tab:builder-stagetime}), and at budget 32 it
raises end-to-end throughput by $22\%$.

\begin{table}[t]
\centering
\footnotesize
\caption{Tree-construction (build) time per round under the Python reference builder and
the fused frontier builder, Qwen3-4B, $T{=}0$, eight datasets (430 rows per configuration).
Both builders select the same nodes. Ratio is Python time over frontier time.}
\label{tab:builder-stagetime}
\resizebox{\columnwidth}{!}{%
\begin{tabular}{lccccc}
\toprule
\multirow{2}{*}{Budget} & \multicolumn{3}{c}{Build (ms per round)} & \multicolumn{2}{c}{Throughput (tok/s)} \\
\cmidrule(lr){2-4}\cmidrule(lr){5-6}
 & Python & Frontier & Ratio & Python & Frontier \\
\midrule
16 & 3.56 & 0.68 & 5.2$\times$ & 331.9 & 378.0 \\
32 & 5.80 & 0.88 & 6.6$\times$ & 331.1 & 404.8 \\
64 & 10.40 & 1.26 & 8.3$\times$ & 266.5 & 355.5 \\
\bottomrule
\end{tabular}}
\end{table}

Table~\ref{tab:builder-compare} compares both builders against Domino's chain run in our own
harness. At budget 16, DominoTree wins at every temperature with either builder, by $5$--$7\%$
with the Python builder and $20$--$21\%$ with the frontier builder: the builder decides how much
the conditional tree wins, not whether. At budget 32 the frontier builder's margin is $28.7\%$
(Table~\ref{tab:budget-ablation}). Whether conditional beats marginal scoring is answered
independently of the builder in Appendix~\ref{app:conditioning} (Table~\ref{tab:c3}).

\begin{table}[t]
\centering
\footnotesize
\caption{DominoTree's throughput advantage (\%, 95\% CI, paired bootstrap over prompts) over
\textbf{Domino (our harness)} (Appendix~\ref{app:ablations}) under each builder, Qwen3-4B, node
budget 16, $M{=}64$, eight datasets, one process per temperature. The chain builds no tree, so
only DominoTree's side changes between builders.}
\label{tab:builder-compare}
\begin{tabular}{lcc}
\toprule
$T$ & Python builder & Fused frontier builder \\
\midrule
$0$   & +5.32 [+4.09, +6.64] & \textbf{+20.21 [+18.85, +21.66]} \\
$0.5$ & +6.14 [+4.58, +7.71] & \textbf{+20.61 [+18.84, +22.38]} \\
$1$   & +6.60 [+4.95, +8.29] & \textbf{+21.24 [+19.37, +23.12]} \\
\bottomrule
\end{tabular}
\end{table}

\subsection{Losslessness}
\label{sec:exp-lossless}

The target verifier is unchanged, so every committed token is one the target itself would emit,
whatever the tree proposed. The only caveat, shared by every tree method that verifies in bf16
(DDTree and SpecInfer included), is that a near-exact tie can break differently from plain AR
decoding (Section~\ref{sec:background-specdec}).

\subsection{Serving on SGLang}
\label{sec:serving}

This section extends the evaluation from the single-stream research harness of Table~\ref{tab:main}
to a production serving engine, in three settings: single-request throughput, \emph{goodput}
under concurrent load, and long context. The plugin, the five methods, the fairness rule and the hardware
are described in Section~\ref{sec:exp-setup}. The served chain shares the drafter, correction head
and engine with DominoTree, so this is the tightest tree-versus-chain comparison in the paper, and a
stronger chain baseline than the HF harness, where DominoTree leads the released decoder by
$19.81$--$26.05\%$ on Qwen3-4B. The served builder is the frontier builder: the sequential heap's
per-pop host synchronization would be paid per request and serialized across the batch, while one
frontier launch covers every in-flight request.

\subsubsection{Single-stream (batch size 1)}
\label{sec:serving-bs1}

Table~\ref{tab:sglang-bs1} compares AR, EAGLE-3\footnote{The Qwen3-4B EAGLE-3 checkpoint
(\texttt{AngelSlim/Qwen3-4B\_eagle3}) declares an \texttt{architectures} class name that this SGLang
build does not register. We changed only that field, to \texttt{LlamaForCausalLMEagle3}, the name
the Qwen3-8B EAGLE-3 checkpoint already uses, and left the weights untouched; a static load trace
confirms every tensor still loads.}, DFlash, the Domino chain and DominoTree on the prompts and
generation cap of Section~\ref{sec:exp-setup} (50 per dataset, 30 for AIME25, up to 2048 tokens).
Here $\tau$ is the mean over prompts of each prompt's accepted length.

DominoTree has the highest accepted length of the five methods in all $32$ model,
temperature and dataset settings, and higher throughput than the chain in all $32$, significantly
in $27$ (Table~\ref{tab:sglang-bs1-cis}). Averaged over the datasets at $T{=}0$, it is $1.17\times$
as fast as the chain on Qwen3-4B and $1.09\times$ on Qwen3-8B, and faster than AR, EAGLE-3 and DFlash
at both sizes.

Less of the acceptance lead turns into throughput at Qwen3-8B. DominoTree accepts about $27\%$
more tokens per round than the chain at both sizes, but its throughput lead at Qwen3-8B is about
half that at Qwen3-4B. Throughput is accepted tokens per round divided by time per round, so the
difference lies in round time: a DominoTree round takes $17\%$ longer than a chain round at Qwen3-8B,
against $10\%$ at Qwen3-4B. The likely cause is verification: DominoTree verifies $33$ tokens per
round against the chain's $16$, which costs more on the larger model.

\begin{table*}[t]
\centering
\footnotesize
\setlength{\tabcolsep}{3pt}
\caption{SGLang serving at batch size 1: speedup over AR and mean accepted length $\tau$, $T{=}0$
and $T{=}1$, one serving session per model. Qwen3-4B and Qwen3-8B run on one RTX~5090, Qwen3.6-27B
on one H200, without tensor parallelism; speedups are normalized to AR on the same machine (AR:
$156$--$168$, $92$--$98$ and $64$--$68$\,tok/s). Tree budget 32 for Qwen3-4B/8B ($33$ verify
slots against the chain's $16$) and 15 for Qwen3.6-27B. \textbf{MTP} is Qwen3.6-27B's built-in
multi-token-prediction head; no EAGLE-3 checkpoint exists for that model. Overall and bold as
in Table~\ref{tab:main}.}

\label{tab:sglang-bs1}
\resizebox{\textwidth}{!}{%
\begin{tabular}{ll*{9}{cc}}
\toprule
Model & Method & \multicolumn{6}{c}{Math} & \multicolumn{6}{c}{Code} & \multicolumn{4}{c}{Chat} & \multicolumn{2}{c}{Overall} \\
\cmidrule(lr){3-8}\cmidrule(lr){9-14}\cmidrule(lr){15-18}\cmidrule(lr){19-20}
 & & \multicolumn{2}{c}{GSM8K} & \multicolumn{2}{c}{MATH-500} & \multicolumn{2}{c}{AIME25} & \multicolumn{2}{c}{HumanEval} & \multicolumn{2}{c}{MBPP} & \multicolumn{2}{c}{LCB} & \multicolumn{2}{c}{MT-Bench} & \multicolumn{2}{c}{Alpaca} & \multicolumn{2}{c}{Avg.} \\
\midrule
\multicolumn{2}{c}{Temperature = 0} & Speedup & $\tau$ & Speedup & $\tau$ & Speedup & $\tau$ & Speedup & $\tau$ & Speedup & $\tau$ & Speedup & $\tau$ & Speedup & $\tau$ & Speedup & $\tau$ & Speedup & $\tau$ \\
\midrule
\multirow{4}{*}{Qwen3-4B}
 & EAGLE-3 & 1.44 & 3.38 & 1.34 & 3.12 & 1.39 & 3.11 & 1.34 & 3.11 & 1.28 & 3.03 & 1.22 & 2.85 & 1.34 & 3.13 & 1.15 & 2.80 & 1.31 & 3.06 \\
 & DFlash & 3.55 & 6.59 & 4.29 & 7.92 & 4.13 & 7.37 & 3.59 & 6.77 & 3.29 & 6.26 & 3.45 & 7.07 & 2.35 & 4.45 & 1.54 & 2.90 & 3.27 & 6.17 \\
 & Domino & 4.94 & 9.89 & 4.55 & 8.81 & 3.98 & 7.45 & 3.52 & 7.04 & 3.62 & 7.22 & 3.11 & 6.40 & 2.64 & 5.20 & 1.96 & 3.90 & 3.54 & 6.99 \\
 & \textbf{DominoTree (32)} & \textbf{5.46} & \textbf{12.02} & \textbf{5.15} & \textbf{11.02} & \textbf{4.67} & \textbf{9.61} & \textbf{4.21} & \textbf{9.01} & \textbf{4.29} & \textbf{9.42} & \textbf{3.55} & \textbf{8.13} & \textbf{3.13} & \textbf{6.81} & \textbf{2.40} & \textbf{5.23} & \textbf{4.11} & \textbf{8.91} \\
\midrule
\multirow{4}{*}{Qwen3-8B}
 & EAGLE-3 & 1.76 & 3.30 & 1.71 & 3.09 & 1.71 & 3.04 & 1.79 & 3.23 & 1.56 & 2.86 & 1.50 & 2.76 & 1.59 & 2.90 & 1.34 & 2.57 & 1.62 & 2.97 \\
 & DFlash & 4.18 & 6.59 & 5.24 & 7.97 & 4.76 & 7.14 & 4.32 & 6.60 & 3.86 & 5.99 & 4.30 & 7.38 & 2.72 & 4.22 & 1.94 & 3.12 & 3.91 & 6.13 \\
 & Domino & 5.99 & 9.86 & 5.69 & 9.09 & 4.94 & 7.58 & 4.69 & 7.44 & 4.38 & 7.00 & 4.31 & 7.57 & 3.15 & 5.08 & 2.49 & 4.07 & 4.46 & 7.21 \\
 & \textbf{DominoTree (32)} & \textbf{6.31} & \textbf{12.02} & \textbf{6.11} & \textbf{11.36} & \textbf{5.36} & \textbf{9.72} & \textbf{5.17} & \textbf{9.65} & \textbf{4.97} & \textbf{9.37} & \textbf{4.37} & \textbf{9.02} & \textbf{3.54} & \textbf{6.76} & \textbf{2.83} & \textbf{5.50} & \textbf{4.83} & \textbf{9.17} \\
\midrule
\multirow{4}{*}{Qwen3.6-27B}
 & MTP & 2.59 & 3.67 & 2.66 & 3.68 & 2.59 & 3.55 & 2.64 & 3.83 & 2.42 & 3.63 & 2.32 & 3.31 & 2.21 & 3.13 & 2.02 & 2.99 & 2.43 & 3.47 \\
 & DFlash & 4.71 & 7.42 & 5.33 & 8.05 & 4.65 & 6.84 & 6.24 & 11.02 & \textbf{4.34} & 7.73 & \textbf{3.41} & 5.63 & 3.02 & 4.74 & 2.18 & 3.50 & 4.23 & 6.87 \\
 & Domino & \textbf{5.81} & 9.41 & 5.99 & 9.20 & 4.82 & 7.22 & \textbf{6.31} & 11.06 & 4.33 & 7.52 & 3.36 & 5.46 & 3.26 & 5.22 & 2.47 & 3.91 & 4.54 & 7.37 \\
 & \textbf{DominoTree (15)} & 5.58 & \textbf{9.77} & \textbf{6.06} & \textbf{9.83} & \textbf{5.28} & \textbf{8.36} & 5.80 & \textbf{11.61} & \textbf{4.34} & \textbf{8.07} & 3.30 & \textbf{6.22} & \textbf{3.45} & \textbf{6.02} & \textbf{2.87} & \textbf{4.75} & \textbf{4.58} & \textbf{8.08} \\
\midrule
\multicolumn{2}{c}{Temperature = 1} & Speedup & $\tau$ & Speedup & $\tau$ & Speedup & $\tau$ & Speedup & $\tau$ & Speedup & $\tau$ & Speedup & $\tau$ & Speedup & $\tau$ & Speedup & $\tau$ & Speedup & $\tau$ \\
\midrule
\multirow{4}{*}{Qwen3-4B}
 & EAGLE-3 & 1.36 & 3.28 & 1.27 & 2.99 & 1.23 & 2.81 & 1.29 & 3.05 & 1.21 & 2.98 & 1.19 & 2.82 & 1.25 & 2.98 & 1.07 & 2.71 & 1.23 & 2.95 \\
 & DFlash & 3.14 & 5.87 & 3.63 & 6.67 & 2.79 & 4.97 & 3.32 & 6.12 & 2.99 & 5.65 & \textbf{3.42} & 7.00 & 2.16 & 4.00 & 1.49 & 2.85 & 2.87 & 5.39 \\
 & Domino & 4.25 & 8.33 & 3.81 & 7.32 & 2.64 & 4.93 & 3.33 & 6.41 & 3.34 & 6.61 & 2.96 & 6.27 & 2.29 & 4.51 & 1.79 & 3.54 & 3.05 & 5.99 \\
 & \textbf{DominoTree (32)} & \textbf{4.48} & \textbf{10.06} & \textbf{4.11} & \textbf{8.96} & \textbf{2.95} & \textbf{6.21} & \textbf{3.85} & \textbf{8.39} & \textbf{3.97} & \textbf{8.87} & 3.34 & \textbf{7.83} & \textbf{2.76} & \textbf{6.15} & \textbf{2.25} & \textbf{5.03} & \textbf{3.46} & \textbf{7.69} \\
\midrule
\multirow{4}{*}{Qwen3-8B}
 & EAGLE-3 & 1.69 & 3.16 & 1.62 & 2.98 & 1.50 & 2.69 & 1.70 & 3.11 & 1.52 & 2.81 & 1.41 & 2.62 & 1.47 & 2.74 & 1.27 & 2.46 & 1.52 & 2.82 \\
 & DFlash & 3.75 & 5.79 & 4.33 & 6.57 & 3.22 & 4.79 & 3.71 & 5.66 & 3.46 & 5.37 & 3.98 & 6.97 & 2.35 & 3.63 & 1.83 & 2.91 & 3.33 & 5.21 \\
 & Domino & 5.11 & 8.27 & 4.61 & 7.23 & 3.14 & 4.81 & 3.84 & 6.06 & 3.70 & 5.89 & 3.97 & 7.04 & 2.65 & 4.25 & 2.18 & 3.53 & 3.65 & 5.88 \\
 & \textbf{DominoTree (32)} & \textbf{5.17} & \textbf{9.91} & \textbf{4.67} & \textbf{8.79} & \textbf{3.45} & \textbf{6.33} & \textbf{4.07} & \textbf{7.69} & \textbf{4.11} & \textbf{7.85} & \textbf{4.05} & \textbf{8.56} & \textbf{2.91} & \textbf{5.62} & \textbf{2.57} & \textbf{5.06} & \textbf{3.87} & \textbf{7.48} \\
\midrule
\multirow{4}{*}{Qwen3.6-27B}
 & MTP & 2.50 & 3.60 & 2.54 & 3.59 & 2.45 & 3.43 & 2.59 & 3.80 & 2.34 & 3.51 & 2.18 & 3.19 & 1.98 & 2.89 & 1.86 & 2.79 & 2.31 & 3.35 \\
 & DFlash & 4.52 & 7.07 & 5.10 & 7.68 & 4.28 & 6.28 & 6.11 & 10.61 & 4.12 & 7.22 & 3.39 & 5.54 & 2.74 & 4.33 & 1.99 & 3.22 & 4.03 & 6.49 \\
 & Domino & 5.50 & 8.74 & 5.50 & 8.39 & 4.50 & 6.67 & \textbf{6.37} & 11.00 & 4.05 & 6.82 & 3.26 & 5.33 & 2.96 & 4.78 & 2.16 & 3.45 & 4.29 & 6.90 \\
 & \textbf{DominoTree (15)} & \textbf{5.52} & \textbf{9.30} & \textbf{5.59} & \textbf{9.01} & \textbf{4.88} & \textbf{7.69} & 6.22 & \textbf{11.32} & \textbf{4.23} & \textbf{7.63} & \textbf{3.54} & \textbf{6.05} & \textbf{3.24} & \textbf{5.42} & \textbf{2.49} & \textbf{4.24} & \textbf{4.46} & \textbf{7.58} \\
\bottomrule
\end{tabular}
}
\end{table*}

\begin{table*}[t]
\centering
\footnotesize
\caption{DominoTree's throughput advantage over the Domino chain at batch size 1 (\%, 95\% CI,
paired bootstrap over prompts), from the runs of Table~\ref{tab:sglang-bs1}. The overall row is
the mean of the eight dataset values. $^{*}$ marks a CI that includes zero ($5$ of $32$
settings); every overall value is significant.}
\label{tab:sglang-bs1-cis}
\resizebox{\textwidth}{!}{%
\begin{tabular}{l*{4}{cc}}
\toprule
\multirow{2}{*}{Dataset} & \multicolumn{2}{c}{Qwen3-4B, $T{=}0$} & \multicolumn{2}{c}{Qwen3-4B, $T{=}1$} & \multicolumn{2}{c}{Qwen3-8B, $T{=}0$} & \multicolumn{2}{c}{Qwen3-8B, $T{=}1$} \\
\cmidrule(lr){2-3}\cmidrule(lr){4-5}\cmidrule(lr){6-7}\cmidrule(lr){8-9}
 & $\Delta\%$ & 95\% CI & $\Delta\%$ & 95\% CI & $\Delta\%$ & 95\% CI & $\Delta\%$ & 95\% CI \\
\midrule
GSM8K & +10.3 & $[+8.2,+12.6]$ & +5.5$^{*}$ & $[-0.2,+11.2]$ & +5.3 & $[+3.3,+7.3]$ & +1.3$^{*}$ & $[-3.4,+6.6]$ \\
MATH-500 & +13.3 & $[+10.3,+16.4]$ & +7.9 & $[+3.6,+12.6]$ & +7.4 & $[+4.6,+10.5]$ & +1.5$^{*}$ & $[-2.4,+6.0]$ \\
AIME25 & +17.2 & $[+12.7,+22.6]$ & +11.6 & $[+6.4,+17.2]$ & +8.6 & $[+1.4,+14.9]$ & +9.7 & $[+3.7,+16.2]$ \\
\cmidrule(lr){1-9}
HumanEval & +19.7 & $[+16.8,+22.6]$ & +15.6 & $[+10.2,+20.6]$ & +10.2 & $[+7.7,+13.1]$ & +5.8 & $[+1.3,+10.6]$ \\
MBPP & +18.7 & $[+15.6,+22.0]$ & +18.7 & $[+13.1,+24.7]$ & +13.3 & $[+10.2,+16.9]$ & +11.0 & $[+6.2,+16.3]$ \\
LCB & +14.2 & $[+10.9,+17.7]$ & +12.8 & $[+8.8,+17.2]$ & +1.4$^{*}$ & $[-4.8,+7.3]$ & +2.0$^{*}$ & $[-1.6,+5.9]$ \\
\cmidrule(lr){1-9}
MT-Bench & +18.7 & $[+15.2,+22.8]$ & +20.4 & $[+15.8,+25.5]$ & +12.4 & $[+8.7,+16.4]$ & +9.7 & $[+4.8,+14.9]$ \\
Alpaca & +22.1 & $[+17.8,+26.6]$ & +25.7 & $[+19.7,+32.0]$ & +13.7 & $[+8.7,+18.5]$ & +17.7 & $[+12.5,+22.7]$ \\
\midrule
\textbf{Overall} & +16.8 & $[+15.6,+18.1]$ & +14.8 & $[+13.0,+16.7]$ & +9.0 & $[+7.5,+10.6]$ & +7.3 & $[+5.7,+9.1]$ \\
\bottomrule
\end{tabular}
}
\end{table*}

\subsubsection{Serving under concurrency}
\label{sec:serving-conc}

We measure goodput as the offered concurrency $c$ rises to $32$ (Table~\ref{tab:sglang-conc}).
Acceptance does not depend on load: each method's $\tau$ changes by less than $0.1$ from $c{=}2$
to $32$, and DominoTree's stays about $28\%$ above the chain's. Goodput does:

\begin{center}\small
\resizebox{\columnwidth}{!}{
\begin{tabular}{lrrrrr}
\toprule
$\Delta$ goodput vs.\ Domino & $c{=}2$ & $c{=}4$ & $c{=}8$ & $c{=}16$ & $c{=}32$ \\
\midrule
Qwen3-4B & $+6.2\%$ & $+0.2\%$ & $-19.3\%$ & $-28.1\%$ & $-26.7\%$ \\
Qwen3-8B & $+8.1\%$ & $-5.3\%$ & $-21.1\%$ & $-21.2\%$ & $-23.8\%$ \\
\bottomrule
\end{tabular}
}
\end{center}

\noindent The tree leads at $c{=}2$; at Qwen3-4B the two are even at $c{=}4$, and at Qwen3-8B the
chain already leads there. A likely mechanism, which we did not instrument: at low load the GPU has
spare compute, so the tree's extra verification is nearly free; at saturation that work competes
with other requests, and the same $+28\%$ acceptance becomes a goodput deficit of about $20\%$. This
defines an operating regime rather than a defect: the tree suits latency-oriented serving
(single stream, long context, light load), the chain saturated throughput. At
saturation speculation itself can hurt: EAGLE-3 falls below AR at $c{=}16$ and $32$ on all three
datasets.

\begin{table*}[t]
\centering
\small
\caption{SGLang goodput (aggregate output tok/s) at offered concurrency $c{=}2$ to $32$, $T{=}0$,
512-token generations, hardware as in Table~\ref{tab:sglang-bs1}. Overall is the unweighted mean
over GSM8K, MBPP and MT-Bench. $\tau$ is averaged over $c$; for Qwen3-4B/8B it varies by
$<0.1$ across $c$. Per-dataset detail is in Tables~\ref{tab:app-conc} and~\ref{tab:app-conc-27b}.
Bold marks the best value per block.}
\label{tab:sglang-conc}
\begin{tabular}{llcccccc}
\toprule
Model & Method & $c{=}2$ & $c{=}4$ & $c{=}8$ & $c{=}16$ & $c{=}32$ & $\tau$ \\
\midrule
\multirow{5}{*}{Qwen3-4B}
 & AR             & 295 & 568 & 972 & 1739 & 3011 & -- \\
 & EAGLE-3        & 390 & 689 & 1084 & 1586 & 1860 & 3.09 \\
 & DFlash         & 866 & 1555 & 2364 & 3318 & 4045 & 5.54 \\
 & Domino         & 1002 & 1731 & \textbf{2652} & \textbf{3672} & \textbf{4569} & 7.21 \\
 & DominoTree (32) & \textbf{1063} & \textbf{1734} & 2141 & 2642 & 3350 & \textbf{9.20} \\
\midrule
\multirow{5}{*}{Qwen3-8B}
 & AR             & 185 & 348 & 652 & 1188 & 1667 & -- \\
 & EAGLE-3        & 247 & 459 & 724 & 977 & 1204 & 2.97 \\
 & DFlash         & 484 & 920 & 1449 & 1820 & 2236 & 5.54 \\
 & Domino         & 604 & \textbf{1123} & \textbf{1751} & \textbf{2212} & \textbf{2753} & 7.27 \\
 & DominoTree (32) & \textbf{653} & 1064 & 1382 & 1742 & 2099 & \textbf{9.30} \\
\midrule
\multirow{5}{*}{Qwen3.6-27B}
 & AR             & 126 & 234 & 410 & 328 & 568 & -- \\
 & MTP            & 296 & 518 & 835 & 784 & 979 & 3.49 \\
 & DFlash         & 444 & 720 & 1030 & 954 & 936 & 6.64 \\
 & Domino         & 496 & 794 & \textbf{1149} & \textbf{1092} & \textbf{1091} & 7.41 \\
 & DominoTree (15) & \textbf{521} & \textbf{819} & 1134 & 1073 & 1068 & \textbf{8.01} \\
\bottomrule
\end{tabular}
\end{table*}

\subsubsection{Long context}
\label{sec:serving-longctx}

Long context stresses verification, whose cost grows with the context while drafting stays nearly
flat. We use the two summarization tasks of HELMET~\citep{helmet2025}, $\infty$Bench-Sum and
Multi-LexSum (the only HELMET tasks that generate enough tokens to measure accepted length), at
8K, 16K and 32K (Table~\ref{tab:sglang-longctx}); every bin exceeds the $3{,}072$ tokens the Domino
drafter was trained on. DominoTree accepts more than the chain in all $12$ settings, each
significantly, and is faster in $11$, significantly in $10$. On Qwen3-4B it is the fastest method
at every length, reaching $1.59\times$ AR at 8K. The exceptions are all at Qwen3-8B and 32K, furthest
outside the drafter's training range, where the gain over the chain is not significant and the
speedup matches EAGLE-3's, whose acceptance holds up best with length. The advantage narrows with
length because prefill, the same for every method, takes a growing share of time, and acceptance
falls for chain and tree alike. Per-task detail is in Appendix~\ref{app:sglang}.

\begin{table*}[t]
\centering
\small
\caption{Long context on HELMET summarization ($\infty$Bench-Sum and Multi-LexSum): mean accepted
length $\tau$ and speedup over AR at batch size 1, $T{=}0$, up to 1200 generated tokens, 50
prompts per task and length, hardware as in Table~\ref{tab:sglang-bs1}. Each entry is the mean
of the two per-task values. Tree budget 16 (Qwen3-4B), 32 (Qwen3-8B) and 15 (Qwen3.6-27B;
Table~\ref{tab:longctx-budget}). Qwen3-8B's DominoTree arm ran in a separate session on the same
machine as its baselines. Bold marks the best value per column, ties included. Per-task detail
and paired CIs are in Table~\ref{tab:app-helmet}.}
\label{tab:sglang-longctx}
\resizebox{\textwidth}{!}{%
\begin{tabular}{l*{18}{c}}
\toprule
 & \multicolumn{6}{c}{Qwen3-4B ($B{=}16$)} & \multicolumn{6}{c}{Qwen3-8B ($B{=}32$)} & \multicolumn{6}{c}{Qwen3.6-27B ($B{=}15$)} \\
\cmidrule(lr){2-7}\cmidrule(lr){8-13}\cmidrule(lr){14-19}
 & \multicolumn{2}{c}{8K} & \multicolumn{2}{c}{16K} & \multicolumn{2}{c}{32K} & \multicolumn{2}{c}{8K} & \multicolumn{2}{c}{16K} & \multicolumn{2}{c}{32K} & \multicolumn{2}{c}{8K} & \multicolumn{2}{c}{16K} & \multicolumn{2}{c}{32K} \\
\cmidrule(lr){2-3}\cmidrule(lr){4-5}\cmidrule(lr){6-7}\cmidrule(lr){8-9}\cmidrule(lr){10-11}\cmidrule(lr){12-13}\cmidrule(lr){14-15}\cmidrule(lr){16-17}\cmidrule(lr){18-19}
Method & $\tau$ & Speedup & $\tau$ & Speedup & $\tau$ & Speedup & $\tau$ & Speedup & $\tau$ & Speedup & $\tau$ & Speedup & $\tau$ & Speedup & $\tau$ & Speedup & $\tau$ & Speedup \\
\midrule
EAGLE-3 & 2.69 & 1.18$\times$ & 2.65 & 1.17$\times$ & \textbf{2.73} & 1.16$\times$ & 2.50 & 1.35$\times$ & 2.47 & 1.26$\times$ & 2.46 & \textbf{1.21$\times$} & -- & -- & -- & -- & -- & -- \\
MTP & -- & -- & -- & -- & -- & -- & -- & -- & -- & -- & -- & -- & 2.76 & 1.85$\times$ & 2.77 & 1.71$\times$ & 2.78 & \textbf{1.51$\times$} \\
DFlash & 2.09 & 1.19$\times$ & 1.81 & 1.05$\times$ & 1.64 & 1.02$\times$ & 2.05 & 1.32$\times$ & 1.71 & 1.13$\times$ & 1.46 & 1.03$\times$ & 2.73 & 1.73$\times$ & 2.71 & 1.60$\times$ & 2.72 & 1.45$\times$ \\
Domino & 2.54 & 1.31$\times$ & 2.22 & 1.17$\times$ & 1.96 & 1.09$\times$ & 2.52 & 1.50$\times$ & 2.28 & 1.32$\times$ & 2.00 & 1.20$\times$ & 2.97 & 1.84$\times$ & 2.78 & 1.62$\times$ & 2.56 & 1.38$\times$ \\
\textbf{DominoTree} & \textbf{3.41} & \textbf{1.59$\times$} & \textbf{3.02} & \textbf{1.38$\times$} & 2.61 & \textbf{1.22$\times$} & \textbf{3.66} & \textbf{1.71$\times$} & \textbf{3.34} & \textbf{1.43$\times$} & \textbf{2.90} & \textbf{1.21$\times$} & \textbf{3.85} & \textbf{2.11$\times$} & \textbf{3.59} & \textbf{1.83$\times$} & \textbf{3.35} & 1.50$\times$ \\
\bottomrule
\end{tabular}
}
\end{table*}

\paragraph{The best tree budget depends on the model and the context length.} A wider tree raises
accepted length in every setting (Table~\ref{tab:longctx-budget}), but each node adds verification
work that grows with the context. At Qwen3-4B and Qwen3.6-27B the narrowest budget is fastest at
every length; at Qwen3-8B budget 32 is fastest up to 16K, the narrowest wins at 32K, and 64 is
slowest throughout. The best budget thus shrinks with context and does not transfer across models:
Qwen3-8B at Qwen3-4B's best budget loses about $7\%$ at 8K. These runs cannot separate the model
from the hardware (RTX~5090 for Qwen3-4B/8B, H200 for Qwen3.6-27B).

\begin{table}[t]
\centering
\small
\caption{DominoTree's long-context throughput (tok/s) by tree budget, mean over the two HELMET
tasks, $T{=}0$, 50 prompts per task and length, hardware as in Table~\ref{tab:sglang-bs1}. Bold
marks the best budget per row. $^{\dagger}$The narrowest budget is 16 for Qwen3-4B/8B and 15 for
Qwen3.6-27B. The Qwen3-4B row ran on a different RTX~5090 host than
Table~\ref{tab:sglang-longctx}, so its absolute throughput is not comparable with that table.}
\label{tab:longctx-budget}
\begin{tabular}{llccc}
\toprule
Model & Context & Narrowest$^{\dagger}$ & $B{=}32$ & $B{=}64$ \\
\midrule
\multirow{3}{*}{Qwen3-4B}
 & 8K  & \textbf{231.4} & 225.5 & 203.5 \\
 & 16K & \textbf{157.0} & 150.2 & 132.4 \\
 & 32K & \textbf{90.7} & 83.7 & 76.0 \\
\midrule
\multirow{3}{*}{Qwen3-8B}
 & 8K  & 126.1 & \textbf{136.0} & 113.9 \\
 & 16K & 90.7 & \textbf{94.6} & 80.7 \\
 & 32K & \textbf{58.6} & 56.6 & 51.1 \\
\midrule
\multirow{3}{*}{Qwen3.6-27B}
 & 8K  & \textbf{137.7} & 127.2 & 105.9 \\
 & 16K & \textbf{105.5} & 98.4 & 83.0 \\
 & 32K & \textbf{72.1} & 66.7 & 57.2 \\
\bottomrule
\end{tabular}
\end{table}

Taken together, DominoTree accepts more than the chain on every axis at both model sizes, and turns
that into a throughput lead for single-request serving, for long context except Qwen3-8B at 32K,
and at light load: it leads at $c{=}2$ on both models, ties the chain at $c{=}4$ on Qwen3-4B and
trails it there on Qwen3-8B. Under saturated load the chain sustains more goodput. Where
these crossovers fall depends on the GPU's verification cost (Section~\ref{sec:limitations}).

\subsubsection{Scaling to a 27B hybrid-attention target}
\label{sec:serving-27b}

Qwen3.6-27B combines $48$ Gated-DeltaNet recurrent layers with $16$ full-attention layers. It runs
on one H200 (141\,GB) without tensor parallelism, on a different machine from the Qwen3-4B/8B
serving runs; speedups are normalized to AR on the same machine. DominoTree and the chain share
the \texttt{Huang2020/qwen3.6-27B-domino} drafter. The baselines are AR, DFlash, the chain and the
target's own multi-token-prediction head (MTP), run with SGLang's NEXTN algorithm and the model
card's settings ($3$ steps, top-$1$, $4$ draft tokens); no EAGLE-3 drafter exists for this target.
Serving flags are matched within each axis.

\paragraph{Tree-aware recurrent verification.} An ancestor-only attention mask verifies a tree in
one pass (Section~\ref{sec:method-verify}), but that is not enough for Gated-DeltaNet: if the tree
is flattened into a chain, a rejected sibling's tokens update the recurrent state that later,
accepted nodes read. The integration instead uses parent-pointer Gated-DeltaNet verification
kernels with a dense convolution window per node, and commits only the accepted path's recurrent
state.

To show the failure is real, we turned tree-aware verification off, so the recurrent state is
updated as if the tree were one chain, and measured \emph{prefix agreement}: how many tokens an
output matches plain AR decoding of the same prompt before the first difference. With it on, the
median prompt matches AR for $306$ tokens, close to the chain's $333$; with it off, for only $46$,
although mean accepted length is unchanged ($9.59$ on, $9.58$ off). Accepted length cannot detect
this: a target running on a corrupted state still produces fluent tokens that the draft keeps
matching, while the text drifts from what AR would write. This first check ran on two H100 GPUs with
tensor parallelism and $M{=}64$; every configuration reported below (one H200, no tensor
parallelism, GRU input table on) passed the same check against AR references regenerated on that
GPU. The check is relative to the chain and the negative control, not a claim of byte-identical
output, since bf16 verification can break near-ties differently from AR.

\paragraph{Retuning the tree budget.} The sweep in Table~\ref{tab:app-qwen27b-ablation} selects
$B{=}15$, $k{=}8$, $M{=}128$ with the GRU input table for every axis. Smaller trees lose
acceptance; larger ones gain acceptance but lose throughput, because verification, not build,
dominates every round and its cost grows faster than the acceptance it buys. $B{=}15$ is also
fastest at all three context lengths (Table~\ref{tab:longctx-budget}). Acceptance is insensitive
to $k$ and $M$, and correcting over the full vocabulary gains nothing and is slower.

\paragraph{Results.} At batch size 1 (Table~\ref{tab:sglang-bs1}), DominoTree has the highest
accepted length in every dataset and temperature but is about tied with the chain on throughput:
$4.58\times$ against $4.54\times$ AR at $T{=}0$ and $4.46\times$ against $4.29\times$ at $T{=}1$; the
chain wins GSM8K and HumanEval at $T{=}0$ and HumanEval at $T{=}1$, and DFlash wins LiveCodeBench at
$T{=}0$. On HELMET, DominoTree reaches $1.43$--$2.13\times$ AR and improves mean per-prompt
throughput over the chain, with paired confidence intervals above zero, in all $6$ settings,
including a small gain on Multi-LexSum at 32K, where MTP is about as fast on both tasks
(Table~\ref{tab:app-helmet}); the aggregate speedup at 32K is $1.50\times$, against MTP's
$1.51\times$ (Table~\ref{tab:sglang-longctx}; the two use the different estimands stated in the
captions). Under concurrency it leads overall goodput by $3$--$5\%$ at $c{=}2$ and $4$ and trails
the chain by $1$--$2\%$ from $c{=}8$ to $32$; MTP is fastest on MT-Bench at $c{=}32$
(Tables~\ref{tab:sglang-conc} and~\ref{tab:app-conc-27b}), while DominoTree keeps the highest
accepted length throughout. The hybrid target thus keeps the acceptance advantage, and whether it
becomes a throughput gain depends on context and load.

\section{Related Work}
\label{sec:related}

\paragraph{Speculative decoding and tree verification.} Speculative decoding, proposed independently
by Leviathan et al.~\citep{leviathan2023speculative} and Chen et al.~\citep{chen2023accelerating},
was extended from a single draft chain to a token tree by SpecInfer~\citep{miao2024specinfer}, which
verifies the whole tree in one target pass with an ancestor-only attention mask, the mechanism this
paper reuses (Section~\ref{sec:method-verify}).

\paragraph{Other draft structures.} A second family removes the separate draft model.
Medusa~\citep{cai2024medusa} adds decoding heads to the target and verifies combinations of their
predictions as a tree; lookahead decoding~\citep{fu2024lookahead} drafts $n$-grams produced by Jacobi
iteration on the target; and REST~\citep{he2024rest} retrieves continuations from a datastore. These
proposals carry no drafter-internal distribution to condition on, so they are complementary to
DominoTree rather than competing with it.

\paragraph{Draft trees on autoregressive and block-diffusion drafters.} The EAGLE
line~\citep{li2024eagle,li2024eagle2,li2025eagle3} builds trees on an autoregressive drafter, whose
branches each re-run the draft model; Domino instead shares one backbone pass across branches
(Section~\ref{sec:background-domino}). On block-diffusion drafters, DDTree~\citep{ringel2026ddtree}
and CaDDTree~\citep{zhang2026caddtree} build trees from DFlash's~\citep{chen2026dflash} marginals
(Sections~\ref{sec:background-ddtree}--\ref{sec:background-caddtree}); DominoTree keeps their
best-first construction and changes the node score.

\paragraph{Closest related work: path-conditioned trees on parallel drafters.} Three other methods
give a parallel drafter's tree branch-specific distributions. JetSpec~\citep{hu2026jetspec} trains a
new causal parallel draft head aligned with the target's factorization and builds a causal tree
with it. TreeFlash~\citep{rheinboldt2026treeflash} adds a lightweight MLP layer to DFlash, fine-tuned
with the drafter, that conditions each position on the drafter's hidden state and the previous
token; it builds a candidate tree from DFlash's distribution, re-scores its nodes with that layer and
selects the final tree from the re-scored distributions, evaluated at batch size 1 only. Both train new parameters (for TreeFlash, 125M on Qwen3-4B), whereas DominoTree
reuses Domino's released head with no training, and that head's GRU state carries the whole path
rather than only the previous token. The distinction is narrow: a purpose-trained causal head is
strictly more expressive than a GRU correction never trained for branching.

DARTree~\citep{li2026dartree}, which appeared one month after the preprint of this
article~\citep{lin2026dominotree}, is the closest. Like DominoTree it is training-free, scores the
tree with Domino's released correction head, and corrects only the 64 marginal-top tokens per depth
(Section~\ref{sec:method-topm}); it keeps all 64 as children of every branch, where we keep the top
8 per node. Its construction follows the pattern of our frontier builder
(Section~\ref{sec:method-frontier}): it scores a fixed-width candidate tree one depth per batched
call and selects the verification tree best-first afterwards. The two differ in width and selection
score. DARTree's default width (12) is independent of its budget (64), and it adds a per-depth
penalty ($\beta{=}{-0.2}$) to each node's cumulative log-probability; we set the width equal to the
budget and select on the unpenalized score, which reproduces the best-first heap's tree exactly
(Appendix~\ref{app:gpu-builder}). DARTree evaluates Qwen3-4B and Qwen3-8B; we also evaluate the
hybrid Qwen3.6-27B, which needs tree-aware recurrent verification (Section~\ref{sec:serving-27b}).
Its SGLang results agree with ours under load: with the budget fixed at 64, its goodput falls below
Domino's from 4 concurrent requests on both workloads it serves, and a hand-set schedule of smaller
budgets and widths per concurrency level keeps it ahead up to 32. Choosing the budget from the load
at run time remains open for both methods (Section~\ref{sec:limitations}).

\section{Limitations and Future Work}
\label{sec:limitations}

\paragraph{Implementation scope.} The frontier builder runs in our research harness and in SGLang;
porting it to vLLM~\citep{kwon2023vllm} or TensorRT-LLM, each with its own scheduler and KV manager,
is future work. The GRU input table is not yet supported under tensor parallelism, where the
embedding is split across GPUs; every frontier-builder result runs on one GPU with the table on.

\paragraph{Adaptive budgeting remains open.} DominoTree uses a fixed node budget. Our port of
CaDDTree's cost-aware rule does not meaningfully improve on it, because the corrected drafter's path
probabilities over-credit acceptance and the rule runs to its cap
(Appendix~\ref{sec:exp-condadaptive}). The estimator fails, not the framework, which leaves two
repairs to try: calibrating the forecast, or replacing it with the acceptance measured in the
previous round.

\paragraph{Operating points.} The budgets of Table~\ref{tab:main} (32 for Qwen3-4B, 128 for
Qwen3-8B, with $M{=}64$ and $k{=}8$) are platform- and model-specific, not per-workload optima: on the
H100 research harness Qwen3-8B is fastest at 128, while the RTX~5090 serving runs use 32. Matching
budgets also does not isolate the score alone, since DDTree and DominoTree use different drafters
and harnesses; Table~\ref{tab:c3} holds those fixed. A joint per-workload sweep of budget and
branching factor across all tree methods remains open.

\paragraph{Serving under saturated load.} On one RTX~5090 up to $c{=}32$, the acceptance advantage
persists but the wider verification loses goodput to the chain at saturation
(Section~\ref{sec:serving-conc}). The crossover and the best budget depend on hardware, model,
context length and workload, so the measured points are not universal thresholds.
DARTree~\citep{li2026dartree} stays ahead of the chain up to 32 requests with a hand-set budget and
width per concurrency level; load-dependent budgets, fixed or chosen at run time, are yet to be tested for DominoTree.

\paragraph{Serving scope.} We measure only SGLang, on single GPUs without tensor parallelism, and
long context only at batch size 1. For Qwen3.6-27B the throughput gains are smaller: DominoTree ties MTP at 32K
and loses goodput to the chain under saturation. The served Domino baseline is the official drafter driven by our plugin
(Section~\ref{sec:exp-setup}); an official SGLang integration has since been merged
upstream~\citep{huang2026dominosglang}, and comparing against it remains future work.

\paragraph{Long-context gains shrink with length.} The Domino drafter is trained to $3{,}072$
tokens, below every HELMET bin. At Qwen3-8B, DominoTree's accepted length falls from $3.66$ at 8K to
$2.90$ at 32K while EAGLE-3 stays near $2.50$ (Table~\ref{tab:sglang-longctx}); acceptance still
exceeds the chain's, but the throughput gain is not significant on either Qwen3-8B/32K task, and on
Qwen3-4B/32K EAGLE-3 accepts more on $\infty$Bench-Sum (Table~\ref{tab:app-helmet}). A
longer-context drafter could help and is yet to be evaluated.

\section{Conclusion}
\label{sec:conclusion}
DominoTree turns Domino's path-dependent correction into a training-free draft tree.
Its acceptance advantage translates into throughput gains when verification cost
remains affordable, while long context and saturated load limit those gains;
tree-aware recurrent verification extends this result to Qwen3.6-27B on one H200.
The best tree budget differs across target models, GPUs, and context lengths, so it
should be tuned for each deployment rather than fixed once.

\section*{Acknowledgments}
\label{sec:acknowledgments}

This work was supported in part by Delta Electronics, Inc., under Contract 114HT907014, as
identified by National Taiwan University. The Qwen3-4B research-harness experiments were run on a
dual-RTX-5080 (16\,GB each) workstation provided by the MIR Lab in the Department of Computer
Science and Information Engineering, National Taiwan University. The Qwen3-8B research-harness
experiments were run on the Nano5 system of the National Center for High-performance Computing
(NCHC) of the National Institutes of Applied Research (NIAR), which the authors thank for
providing computational and storage resources. The authors also thank Delta Electronics, Inc. for
helpful discussions throughout the project.

\section*{Declaration of Generative AI Use}
The research topic, the core idea, and the overall design of the method and
experiments are the authors' own. The authors used Claude
(Anthropic)~\citep{anthropic2026claude} and Codex (OpenAI)~\citep{openai2026codex}
to produce initial drafts of the text of all sections, which the authors revised
throughout; to write the code in the public repository; and to propose some
method components, experiments and analyses, which the authors reviewed,
corrected and approved. The authors verified all results and take full responsibility for the
content of this article.

\section*{Data Availability Statement}
The source code, the scripts that reproduce every table in this article, and the
raw per-prompt measurement records underlying them are publicly available at
\url{https://github.com/slin-zhq/Domino-Tree}. All eight evaluation datasets are
public; each is cited with its source in Section~\ref{sec:exp-setup}.

\bibliographystyle{plainnat}
\bibliography{refs}

\clearpage
\appendix
\renewcommand{\thesection}{\Alph{section}}
\setcounter{section}{0}

\section{Additional ablations}
\label{app:ablations}

Throughout this appendix, \emph{Domino (our harness)} is Domino's single-path chain run by our
own harness on the released drafter (deterministic draft, accept-on-match verification). It differs
from the official decoder of Table~\ref{tab:main}, but it holds drafter, timing and verification
fixed, so each paired $\Delta\%$ varies only the tree.

\subsection{Budget and candidate-width ablations}
\label{sec:exp-ablations}

Both ablations use the headline protocol (Qwen3-4B, $T{=}0$, up to 2048 new tokens, 50 prompts per
dataset over all eight datasets) and vary the two hyperparameters this paper introduces: the node
budget and the candidate width $M$ (Section~\ref{sec:method-topm}).

\begin{table*}[t]
\centering
\small
\caption{Node-budget ablation: DominoTree at budgets 16, 32 and 64 ($M{=}64$, fused frontier
builder) against \textbf{Domino (our harness)} (Appendix~\ref{app:ablations}), Qwen3-4B, $T{=}0$,
eight datasets, all arms in one process. $\Delta\%$ is DominoTree's throughput advantage with a
95\% CI (paired bootstrap over prompts). Bold marks the best value in each column: lowest build time, highest $\tau$, throughput and $\Delta\%$.}
\label{tab:budget-ablation}
\begin{tabular}{lcccc}
\toprule
Node budget & Build (ms) & $\tau$ & Throughput (tok/s) & $\Delta\%$ vs.\ chain (95\% CI) \\
\midrule
16 & \textbf{0.68} & 7.73 & 378.0 & +20.21 [+18.85, +21.66] \\
32 & 0.88 & 8.69 & \textbf{404.8} & \textbf{+28.73 [+27.14, +30.34]} \\
64 & 1.26 & \textbf{9.40} & 355.5 & +13.06 [+11.45, +14.73] \\
\bottomrule
\end{tabular}
\end{table*}

\paragraph{Node budget.} Table~\ref{tab:budget-ablation} compares node budgets of 16, 32 and 64.
Every budget beats the chain, with confidence intervals excluding zero, and 32 is best ($+28.7\%$,
against $+20.2\%$ at 16 and $+13.1\%$ at 64). The ordering shows the trade-off behind the budget. A
larger tree accepts more tokens ($\tau$ rises from $7.73$ to $8.69$ to $9.40$), but it also makes
verification more expensive ($16.8$\,ms per round at budget 32, $21.3$\,ms at 64), and from 32 to 64
the extra verification costs more than the extra acceptance gains. Building the tree is not what
limits the budget: even at 64 it takes only $1.26$\,ms. A faster builder therefore makes more budgets
worthwhile but does not move the best one.

\begin{table*}[t]
\centering
\small
\caption{Candidate-width ablation: the correction restricted to the top $M\in\{32,64,128\}$
candidates or applied over the full vocabulary, at node budgets 16 and 32, against
\textbf{Domino (our harness)}; Qwen3-4B, $T{=}0$, eight datasets, the RTX~5080 machine of
Table~\ref{tab:main}. The chain was re-run beside each $M$ on the same GPU, and each comparison
pairs runs from one GPU. $\Delta\%$ as in Table~\ref{tab:budget-ablation}. Bold marks the best
value per column within each budget. Table~\ref{tab:topm-saturation} extends the accepted-length comparison to more widths.}
\label{tab:topm-ablation}
\begin{tabular}{llcccc}
\toprule
Budget & Width $M$ & Build (ms) & $\tau$ & Throughput (tok/s) & $\Delta\%$ vs.\ Domino (our harness) (95\% CI) \\
\midrule
\multirow{4}{*}{16} & 32 & \textbf{0.69} & 7.67 & 372.2 & +19.55 [+18.20, +20.94] \\
 & 64 & 0.71 & 7.73 & \textbf{372.6} & +20.10 [+18.75, +21.55] \\
 & 128 & 0.76 & \textbf{7.76} & 371.1 & \textbf{+20.89 [+19.39, +22.50]} \\
 & full vocab. & 5.49 & \textbf{7.76} & 299.7 & $-$1.78 [$-$3.11, $-$0.39] \\
\midrule
\multirow{4}{*}{32} & 32 & \textbf{0.88} & 8.67 & \textbf{400.4} & +28.61 [+27.01, +30.27] \\
 & 64 & 0.91 & 8.69 & 397.9 & +28.25 [+26.54, +29.97] \\
 & 128 & 0.97 & 8.72 & 395.5 & \textbf{+28.82 [+27.10, +30.67]} \\
 & full vocab. & 5.78 & \textbf{8.78} & 327.5 & +7.33 [+5.78, +8.95] \\
\bottomrule
\end{tabular}
\end{table*}

\paragraph{Candidate width.} Table~\ref{tab:topm-ablation} varies how many marginal-top candidates
$M$ the correction is applied to. Accepted length barely changes: at budget 32, $M = 32, 64, 128$ give
$\tau = 8.67, 8.69, 8.72$, and the margin over the chain stays between $+28.3\%$ and $+28.8\%$, while
build time grows slightly ($0.88$ to $0.97$\,ms). Applying the correction to the full vocabulary
instead is a clear loss: it adds at most $0.09$ accepted tokens but makes building 6--8 times slower
($5.5$--$5.8$\,ms), so throughput falls $18$--$20\%$ relative to $M{=}64$, and at budget 16 the
full-vocabulary variant is even slightly slower than the chain ($-1.8\%$). This suggests
that the correction mostly re-ranks tokens the marginal already considers likely; it rarely promotes
a token from outside the marginal top 64 into a node's top 8. We therefore fix $M{=}64$ for Qwen3-4B
and Qwen3-8B (Qwen3.6-27B uses $M{=}128$, Section~\ref{sec:serving-27b}).
Table~\ref{tab:topm-saturation} widens $M$ further, from 16 up to the full vocabulary: accepted
length changes by less than $1\%$ on GSM8K, about $3\%$ on HumanEval and about $5\%$ on Alpaca, with
no consistent direction. Widening $M$ beyond 64 therefore brings no measurable benefit, although these
measurements cannot rule out that a token outside the top 16 occasionally matters.

\begin{table*}[ht]
\centering
\small
\caption{Accepted length $\tau$ at node budget 32 (the setting of Table~\ref{tab:main}) as $M$ grows
from 16 to 1024 and to the full vocabulary, Qwen3-4B, $T{=}0$. $\tau$ depends only on the tree's
nodes, so it holds for every builder that selects the same nodes (Appendix~\ref{app:gpu-builder}). Table~\ref{tab:topm-ablation} reports the corresponding build time and throughput.}
\label{tab:topm-saturation}
\footnotesize
\begin{tabular}{lcccccccc}
\toprule
Dataset & $M{=}16$ & $M{=}32$ & $M{=}64$ & $M{=}128$ & $M{=}256$ & $M{=}512$ & $M{=}1024$ & full vocab \\
\midrule
GSM8K      & 11.96 & 12.04 & 12.01 & 12.06 & 12.01 & 12.01 & 11.98 & 12.07 \\
HumanEval  & 8.83  & 8.81  & 8.93  & 9.02  & 9.01  & 9.02  & 9.02  & 9.09 \\
Alpaca     & 5.05  & 5.20  & 5.25  & 5.28  & 5.39  & 5.25  & 5.28  & 5.31 \\
\bottomrule
\end{tabular}
\end{table*}

\subsection{Draft-sampling ablation}
\label{sec:exp-draftsample}

Our harness drafts deterministically (each node proposes the drafter's most likely tokens) and
samples only the target, whereas the released Domino decoder also samples its draft. Both are
lossless, because a draft token is accepted only if it matches the token the target samples.
Sampling the draft lowers throughput in every configuration we test (Table~\ref{tab:draftsample}).
It lengthens tree construction by $1.0$--$1.7$\,ms per round of about $25$\,ms, and it can lower
acceptance: a proposed token is accepted with the target's probability of that token, and a
sampled draft sometimes proposes a token the drafter itself ranks lower. The chain's accepted
length falls by up to $15.7\%$ (Alpaca, $T{=}1.0$). The tree's moves only by $-2.9\%$ to $+1.3\%$,
which we attribute to each node sampling several children without replacement rather than a
single token. Its throughput loss comes mainly from the longer construction, which costs more than
the change in acceptance in five of the six settings. We therefore draft deterministically
throughout; the SGLang plugin has no sampled-draft path.

\begin{table}[ht]
\centering
\small
\caption{Effect of temperature-sampling the draft instead of drafting deterministically:
$\Delta\%=(\text{sampled}-\text{greedy})/\text{greedy}$ in accepted length and throughput,
Qwen3-4B, node budget 16, with the Python reference builder (the only builder that samples
drafts).}
\label{tab:draftsample}
\begin{tabular}{llcccc}
\toprule
& & \multicolumn{2}{c}{Domino (our harness)} & \multicolumn{2}{c}{DominoTree (16)} \\
\cmidrule(lr){3-4}\cmidrule(lr){5-6}
Dataset & $T$ & $\Delta\tau\%$ & $\Delta$TPS\% & $\Delta\tau\%$ & $\Delta$TPS\% \\
\midrule
\multirow{2}{*}{GSM8K}     & 0.5 & $-3.8$  & $-9.1$  & $-2.1$ & $-5.3$ \\
                           & 1.0 & $-3.6$  & $-6.6$  & $-2.9$ & $-5.4$ \\
\cmidrule(lr){1-6}
\multirow{2}{*}{HumanEval} & 0.5 & $-0.4$  & $-5.1$  & $-0.1$ & $-3.4$ \\
                           & 1.0 & $-7.9$  & $-10.8$ & $+0.6$ & $-3.1$ \\
\cmidrule(lr){1-6}
\multirow{2}{*}{Alpaca}    & 0.5 & $-9.9$  & $-13.9$ & $-2.1$ & $-4.8$ \\
                           & 1.0 & $-15.7$ & $-18.1$ & $+1.3$ & $-2.0$ \\

\bottomrule
\end{tabular}
\end{table}

\subsection{The adaptive-budget variant: a calibration-driven negative result}
\label{sec:exp-condadaptive}
\label{sec:method-condadaptive}

CaDDTree's rule (Section~\ref{sec:background-caddtree}) picks the budget maximizing
$\theta(n) = (1+\Phi(n))/(C_d + C_v(n))$ with $\Phi(n) = \sum_{s\le n}\pi_s$. The identity does not
reference factorization; it sums whatever path probabilities the heap produces, and our builder
already produces $\pi_s = \exp(\ell_s^{\mathrm{cond}})$ for every popped node. The adaptive-budget
variant therefore adds the rule to the pop loop at $O(1)$ cost per pop: run
Algorithm~\ref{alg:dominotree} toward a generous cap, accumulate $\Phi(n)$, and stop when $\theta(n)$
stops improving.

The rule is only as good as $\pi_s$, the drafter's estimate that node $s$ will be accepted. Adding
node $n$ raises $\theta$ only if $\pi_n > \theta(n{-}1)\,[C_v(n)-C_v(n{-}1)]$: its forecast
acceptance must exceed the tokens the extra verification time would yield at the current rate. If
$\pi_s$ overstates acceptance, this test can keep passing until the tree reaches the cap.

\begin{figure}[t]
\centering
\IfFileExists{figures/reliability.pdf}{%
  \includegraphics[width=0.7\linewidth]{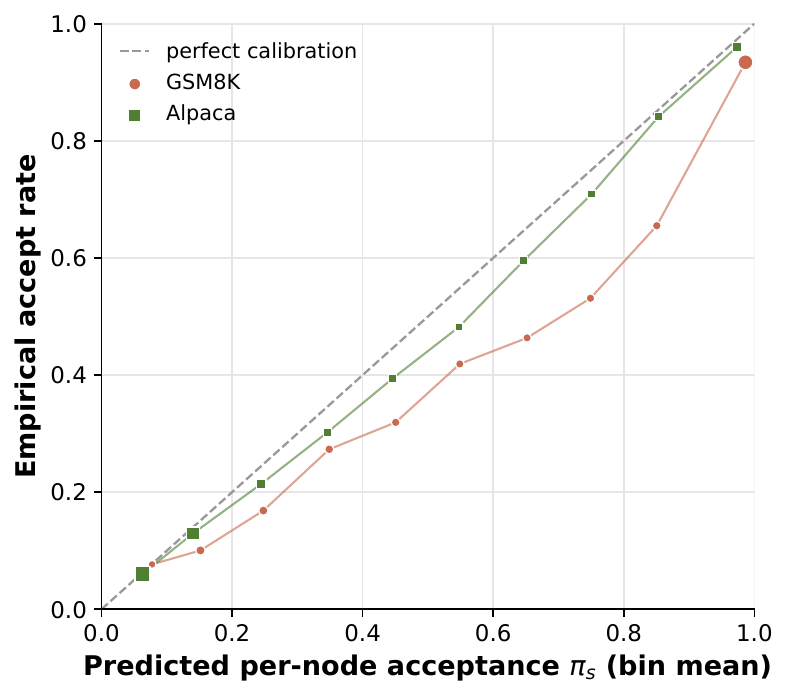}%
}{%
  \fbox{%
    \parbox[c][1.6in][c]{0.68\linewidth}{%
      \centering
      \textbf{Figure placeholder (reliability diagram).}\\[0.4em]
      Predicted per-node acceptance $\pi_s$ vs.\ empirical accept rate,
      GSM8K and Alpaca.
    }%
  }%
}
\caption{Calibration of the adaptive-budget variant: predicted per-node acceptance $\pi_s$ (bin
mean) vs.\ empirical accept rate, GSM8K and Alpaca. The $y{=}x$ line is perfect calibration;
marker area grows with the number of nodes in the bin.}
\label{fig:reliability}
\end{figure}

On the corrected tree, $\pi_s$ is \emph{over-confident}. In a calibration run
($n{=}40$ prompts, $T{=}0$, budget 16, $M{=}64$), the forecast $\Phi$ exceeds the realized accepted
length by $1.16\times$ on GSM8K ($10.73$ against $9.28$) and $1.07\times$ on Alpaca ($3.97$ against
$3.72$), and in every probability bin of Figure~\ref{fig:reliability} the measured acceptance is below
the forecast, more so on math than on chat.

The adaptive variant therefore behaves like a fixed budget equal to its cap. Measured against a fixed
budget of 16 on the same prompts, it gains $+0.5\%$ $[+0.1,+1.2]$ throughput on GSM8K and $+1.4\%$
$[+0.2,+3.8]$ on Alpaca ($n{=}50$ each, paired, $95\%$ CI): measurably above zero, but far too small to
justify the machinery and an order of magnitude below what the tree itself contributes. We use a
fixed budget and report this as a negative result with a known cause: CaDDTree's stopping rule
carries over, but the probability estimate it relies on is over-confident on a corrected drafter.
Calibrating that estimate is left to future work (Section~\ref{sec:limitations}).

\section{Frontier builder: ledger, selection and equivalence}
\label{app:gpu-builder}

\paragraph{Frontier ledger and selection.} At each depth the builder scores all $W{\cdot}k$
children of the frontier, keeps the top $W$ per request and advances the GRU state only for those.
Every scored candidate is written to a ledger of shape $[\,\text{batch},\,D,\,W{\cdot}k\,]$ with its
cumulative log-probability and parent pointer. After $D$ depths the tree is selected in one step:
take the ledger's global top-$B$ entries, order them so parents precede children, remap the parent
pointers, pad short trees with dead leaves, and emit the draft tokens and the ancestor mask. The
frontier scores $D{\cdot}W$ lanes per request where the heap touches $B$ nodes: more arithmetic,
traded for no host synchronization.

\paragraph{Equivalence of the frontier builder to the heap-built tree.} The node score is monotone
along a path: a child's cumulative log-probability is at most its parent's. Best-first expansion
therefore pops exactly the $B$ highest-scoring nodes of the candidate tree, and that set is
ancestor-closed, hence a tree. With $W = B$, fewer than $B$ candidates at any depth can outscore a
top-$B$ node, so none is evicted before its children are scored, and the global top-$B$ of the ledger
is exactly the best-first tree. The one exception is an exact tie, which the heap breaks by insertion
order and the frontier by depth and lane; both results are valid best-first trees, acceptance is
statistically unchanged, and losslessness is unaffected because the verifier is unmodified.
Tests in the public repository check this. On random tie-free inputs at batch sizes 1, 2, 5 and
32, the two builders produce identical trees. With the released drafter weights, on random round
inputs, they agree at budgets 16 and 32; at 64 and 128 they differ in 11 of 16 test rounds, each
time between candidates whose scores are closer than the bf16 accumulation error, so both trees
are valid best-first trees. On GSM8K and Alpaca, accepted length with the frontier builder is
within $0.7\%$ of the heap's at every budget from 16 to 128 ($20$ prompts each). A frontier narrower than the budget ($W<B$) is possible in the code but never used: it could evict a
node of the true best-first tree before its children are scored, drafting a different tree, which
can change $\tau$ but not correctness. A frontier wider than $B$ selects the same tree, up to ties,
at more cost.

\paragraph{Exactness of the fused kernels and the GRU input table.} The fused \texttt{score\_depth}
and \texttt{select\_depth} kernels compute the same quantities as the PyTorch code they replace but
sum in a different order, so a score can differ in its last bit and two candidates tied to within
rounding can swap, as above. We therefore compared accepted length with and without the fused
kernels, and with and without the table-based GRU update, on the benchmark, against a tolerance of
$\pm0.5\%$ fixed in advance: the fused kernels
moved pooled accepted length by $+0.27\%$ (per dataset $-0.31\%$ to $+0.89\%$), and enabling the GRU
table changed nothing at the reported precision. The table costs under one gigabyte of memory at Qwen3's shapes and
is not yet supported under tensor parallelism, where the embedding is split across GPUs.

\begin{table*}[t]
\centering
\small
\caption{DominoTree (budget 32 on 4B, 128 on 8B) against both Domino execution variants: overall
speedup over AR (mean over the eight datasets) and DominoTree's $\Delta\%$ over each.
\emph{Domino-graph} (CUDA graphs) is the stronger variant and the main-body baseline;
\emph{Domino-eager} runs eager PyTorch. Both decode the same chain, so accepted length is omitted.}
\label{tab:domino-variants}
\begin{tabular}{llcccc}
\toprule
Model & $T$ & DominoTree & Domino-graph & Domino-eager & $\Delta\%$ vs.\ graph / eager \\
\midrule
\multirow{3}{*}{Qwen3-4B}
 & 0.0 & 5.33 & 4.45 & 4.28 & \textbf{+19.8} / \textbf{+24.6} \\
 & 0.5 & 5.00 & 4.13 & 3.64 & \textbf{+21.2} / \textbf{+37.5} \\
 & 1.0 & 4.61 & 3.65 & 3.12 & \textbf{+26.1} / \textbf{+47.9} \\
\midrule
\multirow{3}{*}{Qwen3-8B}
 & 0.0 & 6.89 & 5.46 & 5.00 & \textbf{+26.4} / \textbf{+37.8} \\
 & 0.5 & 6.50 & 4.93 & 4.20 & \textbf{+31.8} / \textbf{+54.8} \\
 & 1.0 & 5.75 & 4.25 & 3.41 & \textbf{+35.3} / \textbf{+68.6} \\
\bottomrule
\end{tabular}
\end{table*}

\begin{table*}[t]
\centering
\footnotesize
\caption{Conditioning ladder: accepted length $\tau$ at node budget 16, $M{=}64$, $T{=}0$.
\emph{Marg} scores nodes with the drafter's marginals, \emph{Cond-static} applies Domino's
correction once per depth, and DominoTree recomputes it along each path. \emph{Correction} is
Cond-static over Marg and \emph{Path} is DominoTree over Cond-static, as \% change in $\tau$.
Only $\tau$ is reported: it depends on the tree, not on the code that builds it
(Appendix~\ref{app:conditioning}). $n$ counts prompts (MT-Bench per turn); rollups are
unweighted means over datasets.}
\label{tab:c3}
\resizebox{\textwidth}{!}{%
\begin{tabular}{lc ccccc ccccc}
\toprule
& & \multicolumn{5}{c}{Qwen3-4B} & \multicolumn{5}{c}{Qwen3-8B} \\
\cmidrule(lr){3-7}\cmidrule(lr){8-12}
& & \multicolumn{3}{c}{$\tau$} & \multicolumn{2}{c}{$\Delta\tau$ (\%)} & \multicolumn{3}{c}{$\tau$} & \multicolumn{2}{c}{$\Delta\tau$ (\%)} \\
\cmidrule(lr){3-5}\cmidrule(lr){6-7}\cmidrule(lr){8-10}\cmidrule(lr){11-12}
Dataset / Rollup & $n$ & Marg & Cond-static & DominoTree & Correction & Path & Marg & Cond-static & DominoTree & Correction & Path \\
\midrule
GSM8K & 50 & 9.35 & 10.38 & 10.72 & +11.0 & +3.3 & 9.31 & 10.20 & 10.53 & +9.6 & +3.3 \\
MATH-500 & 50 & 8.68 & 9.44 & 9.77 & +8.8 & +3.5 & 8.78 & 9.80 & 10.09 & +11.5 & +3.0 \\
AIME25 & 30 & 7.26 & 8.13 & 8.48 & +12.0 & +4.3 & 7.64 & 8.05 & 8.47 & +5.4 & +5.2 \\
HumanEval & 50 & 6.91 & 7.59 & 8.00 & +9.9 & +5.4 & 7.17 & 8.06 & 8.47 & +12.4 & +5.1 \\
MBPP & 50 & 6.69 & 7.72 & 8.16 & +15.4 & +5.6 & 6.58 & 7.55 & 7.99 & +14.8 & +5.8 \\
LiveCodeBench & 50 & 7.14 & 7.49 & 7.82 & +5.0 & +4.4 & 7.01 & 7.92 & 8.13 & +13.0 & +2.7 \\
MT-Bench & 100 & 5.30 & 5.78 & 6.14 & +9.1 & +6.1 & 5.25 & 5.72 & 6.01 & +9.0 & +5.0 \\
Alpaca & 50 & 4.06 & 4.44 & 4.78 & +9.6 & +7.5 & 4.07 & 4.50 & 4.83 & +10.7 & +7.3 \\
\midrule
Math & 130 & 8.43 & 9.32 & 9.66 & +10.5 & +3.7 & 8.58 & 9.35 & 9.70 & +9.0 & +3.7 \\
Code & 150 & 6.91 & 7.60 & 7.99 & +10.0 & +5.1 & 6.92 & 7.84 & 8.20 & +13.3 & +4.5 \\
Chat & 150 & 4.68 & 5.11 & 5.46 & +9.3 & +6.7 & 4.66 & 5.11 & 5.42 & +9.8 & +6.0 \\
\midrule
\textbf{Overall} & 430 & 6.92 & 7.62 & 7.98 & +10.1 & +4.7 & 6.98 & 7.73 & 8.07 & +10.7 & +4.4 \\
\bottomrule
\end{tabular}}
\end{table*}

\section{DominoTree vs.\ both Domino execution variants}
\label{app:domino-variants}

The main body uses Domino's stronger execution, the better of its CUDA-graph and eager runners,
which is the CUDA-graph runner in nearly every setting. Table~\ref{tab:domino-variants} reports
DominoTree against each separately. Both decode the same chain, so they differ only in execution
cost. DominoTree beats the CUDA-graph variant at every temperature on both models, and the eager
variant by a much larger margin.

\section{Decomposing the gain: tree structure and conditional scoring}
\label{app:conditioning}
\label{sec:exp-c3}

Relative to a factorized tree, DominoTree changes two things: it applies Domino's correction to the
node score, and it recomputes that correction along each candidate's own path. A single comparison
with a marginal tree measures both together. A third arm that applies the correction without the
path dependence separates them.

\paragraph{The three arms.} All three build a best-first tree at node budget 16 on the same released
Domino drafter, in our harness, with the same verifier, and differ only in the children function
(Section~\ref{sec:method-heap}):
\begin{itemize}
\item \textbf{Marg@16} scores with the base logits alone; every node at a depth shares one menu
  (\emph{factorized}).
\item \textbf{Cond-static@16} applies the correction once per depth, along the single greedy path
  Domino's own decoder walks, and shares that corrected menu across the depth (corrected but still
  \emph{path-independent}).
\item \textbf{DominoTree} recomputes the correction for each node from its own GRU state
  (\emph{path-dependent}).
\end{itemize}
Marg@16 $\to$ Cond-static@16 measures what the correction is worth; Cond-static@16 $\to$ DominoTree
measures what recomputing it per path adds. The second step is measured against this particular
path-independent control, and a different one could split the gain somewhat differently. We use
Marg@16 rather than DDTree as the factorized control because it shares DominoTree's drafter, budget,
harness and verifier; DDTree's rule on Domino's drafter is no longer DDTree as published, so the arm
is named by what it computes. The other contrast, a single chain against a tree when both use Domino's per-path correction,
is the comparison with Domino (our harness) in Tables~\ref{tab:budget-ablation}--\ref{tab:topm-ablation}.

\paragraph{Protocol.} At each model size the three arms run in one session over the same prompts in
the same order (50 per dataset, 30 for AIME25, both MT-Bench turns; $T{=}0$; up to 2048 new tokens),
Qwen3-4B on an RTX~5080 and Qwen3-8B on the H100 of Table~\ref{tab:main}. We compare accepted length
only. It depends only on which nodes a tree contains, not on the code that builds it (the Qwen3-4B
ladder gives identical $\tau$ under two builder implementations), whereas throughput would also
measure how optimized each arm's code is, and only DominoTree has the fused frontier builder.
DominoTree's throughput is in Table~\ref{tab:main} and its builder cost in
Tables~\ref{tab:builder-stagetime} and~\ref{tab:builder-compare}.

\paragraph{Acceptance separates cleanly at both model sizes.} On Qwen3-4B, accepted length rises
$6.92 \to 7.62 \to 7.98$ across the three arms: the correction is worth $+10.1\%$ and recomputing it
per path a further $+4.7\%$, $+15.3\%$ in total, so roughly two thirds of the gain comes from having
the correction and one third from conditioning on the path. Qwen3-8B splits almost the same way,
$6.98 \to 7.73 \to 8.07$ ($+10.7\%$, then $+4.4\%$). Both steps are positive on every dataset at both
sizes (Table~\ref{tab:c3}).

\section{SGLang serving: per-dataset and per-task detail}
\label{app:sglang}
\label{app:qwen27b}

Table~\ref{tab:app-helmet} gives per-task HELMET detail for all three models,
Tables~\ref{tab:app-conc} and~\ref{tab:app-conc-27b} per-dataset concurrency detail, and
Table~\ref{tab:app-qwen27b-ablation} the Qwen3.6-27B tuning sweep behind
Section~\ref{sec:serving-27b}; the full batch-size-1 grid is Table~\ref{tab:sglang-bs1}.

\begin{table*}[p]
\centering\scriptsize\renewcommand{\arraystretch}{0.96}
\caption{HELMET per task and length for all three models, settings as in
Table~\ref{tab:sglang-longctx}. TPS is total output tokens over total time; $\tau$ is mean
per-prompt accepted length. \emph{vs.\ Domino} rows give the paired-bootstrap change against the
chain [95\% CI] in mean per-prompt TPS, a different estimand from the aggregate TPS above them;
$^{*}$ marks a CI containing zero. Qwen3-8B shows budgets 16 and 32
(Table~\ref{tab:sglang-longctx} uses 32). Qwen3.6-27B runs on one H200 with tree budget 15, and
\textbf{MTP} takes EAGLE-3's place, since no EAGLE-3 checkpoint exists for it. DDTree and CaDDTree have no SGLang implementation.
Bold marks the best TPS and $\tau$ per column within a task.}
\label{tab:app-helmet}
\setlength{\tabcolsep}{3pt}
\begin{tabular*}{\textwidth}{@{\extracolsep{\fill}}ll*{3}{cc}}
\toprule
Task & Method & \multicolumn{2}{c}{8K} & \multicolumn{2}{c}{16K} & \multicolumn{2}{c}{32K} \\
\cmidrule(lr){3-4}\cmidrule(lr){5-6}\cmidrule(lr){7-8}
 & & TPS & $\tau$ & TPS & $\tau$ & TPS & $\tau$ \\
\midrule
\multicolumn{8}{c}{\textbf{Qwen3-4B}} \\
\midrule
$\infty$Bench-Sum & AR & 135.7 & 1.00 & 112.4 & 1.00 & 74.1 & 1.00 \\
 & EAGLE-3 & 170.9 & 2.84 & 138.0 & 2.76 & 91.6 & \textbf{2.87} \\
 & DFlash & 165.2 & 2.10 & 118.3 & 1.79 & 77.6 & 1.65 \\
 & Domino & 183.4 & 2.57 & 132.6 & 2.21 & 83.0 & 1.93 \\
 & DominoTree (16) & \textbf{235.4} & \textbf{3.55} & \textbf{162.7} & \textbf{3.03} & \textbf{93.3} & 2.58 \\
 & \quad\emph{vs.\ Domino}, $\Delta$TPS\% & \multicolumn{2}{c}{+29.9 [+24.7,+35.4]} & \multicolumn{2}{c}{+21.1 [+17.8,+24.7]} & \multicolumn{2}{c}{+12.1 [+7.9,+16.3]} \\
 & \quad\emph{vs.\ Domino}, $\Delta\tau$\% & \multicolumn{2}{c}{+38.2 [+32.8,+44.2]} & \multicolumn{2}{c}{+36.9 [+33.0,+41.2]} & \multicolumn{2}{c}{+34.1 [+30.2,+38.0]} \\
\cmidrule(lr){1-8}
Multi-LexSum & AR & 122.3 & 1.00 & 93.3 & 1.00 & 64.3 & 1.00 \\
 & EAGLE-3 & 133.5 & 2.54 & 103.2 & 2.54 & 70.1 & 2.59 \\
 & DFlash & 141.5 & 2.08 & 98.4 & 1.82 & 64.3 & 1.62 \\
 & Domino & 156.1 & 2.52 & 107.6 & 2.23 & 68.6 & 1.99 \\
 & DominoTree (16) & \textbf{177.3} & \textbf{3.28} & \textbf{122.8} & \textbf{3.01} & \textbf{76.3} & \textbf{2.64} \\
 & \quad\emph{vs.\ Domino}, $\Delta$TPS\% & \multicolumn{2}{c}{+13.9 [+11.8,+16.1]} & \multicolumn{2}{c}{+13.7 [+11.3,+16.2]} & \multicolumn{2}{c}{+11.4 [+9.2,+13.6]} \\
 & \quad\emph{vs.\ Domino}, $\Delta\tau$\% & \multicolumn{2}{c}{+30.2 [+27.8,+32.7]} & \multicolumn{2}{c}{+34.8 [+32.1,+37.4]} & \multicolumn{2}{c}{+33.0 [+31.4,+34.7]} \\
\midrule
\multicolumn{8}{c}{\textbf{Qwen3-8B}} \\
\midrule
$\infty$Bench-Sum & AR & 84.6 & 1.00 & 72.7 & 1.00 & 51.3 & 1.00 \\
 & EAGLE-3 & 120.7 & 2.54 & 95.3 & 2.45 & 63.6 & 2.41 \\
 & DFlash & 110.7 & 1.98 & 82.3 & 1.68 & 52.9 & 1.44 \\
 & Domino & 128.2 & 2.45 & 96.7 & 2.22 & 62.1 & 1.93 \\
 & DominoTree (16) & 137.5 & 3.31 & 102.6 & 2.96 & \textbf{66.0} & 2.64 \\
 & \quad\emph{vs.\ Domino}, $\Delta$TPS\% & \multicolumn{2}{c}{+8.8 [+6.6,+11.1]} & \multicolumn{2}{c}{+6.6 [+4.5,+8.8]} & \multicolumn{2}{c}{+6.3 [+3.3,+9.7]} \\
 & \quad\emph{vs.\ Domino}, $\Delta\tau$\% & \multicolumn{2}{c}{+35.1 [+32.4,+37.8]} & \multicolumn{2}{c}{+33.4 [+31.7,+35.3]} & \multicolumn{2}{c}{+37.0 [+34.9,+39.4]} \\
 & DominoTree (32) & \textbf{148.2} & \textbf{3.59} & \textbf{108.9} & \textbf{3.36} & 63.3 & \textbf{2.82} \\
 & \quad\emph{vs.\ Domino}, $\Delta$TPS\% & \multicolumn{2}{c}{+17.3 [+13.2,+22.2]} & \multicolumn{2}{c}{+14.1 [+9.2,+21.4]} & \multicolumn{2}{c}{+1.9 [-1.1,+5.1]$^{*}$} \\
 & \quad\emph{vs.\ Domino}, $\Delta\tau$\% & \multicolumn{2}{c}{+46.4 [+41.5,+52.4]} & \multicolumn{2}{c}{+51.3 [+45.8,+60.0]} & \multicolumn{2}{c}{+46.6 [+43.6,+49.5]} \\
\cmidrule(lr){1-8}
Multi-LexSum & AR & 74.3 & 1.00 & 58.5 & 1.00 & 42.4 & 1.00 \\
 & EAGLE-3 & 94.9 & 2.47 & 71.2 & 2.49 & 50.1 & 2.52 \\
 & DFlash & 99.2 & 2.12 & 66.1 & 1.75 & 43.5 & 1.48 \\
 & Domino & 110.7 & 2.59 & 76.3 & 2.34 & 50.2 & 2.07 \\
 & DominoTree (16) & 114.7 & 3.38 & 78.9 & 3.08 & \textbf{51.3} & 2.74 \\
 & \quad\emph{vs.\ Domino}, $\Delta$TPS\% & \multicolumn{2}{c}{+3.5 [+2.2,+5.0]} & \multicolumn{2}{c}{+3.4 [+2.0,+4.8]} & \multicolumn{2}{c}{+2.5 [+0.4,+4.7]} \\
 & \quad\emph{vs.\ Domino}, $\Delta\tau$\% & \multicolumn{2}{c}{+30.4 [+28.3,+32.8]} & \multicolumn{2}{c}{+31.5 [+29.4,+33.5]} & \multicolumn{2}{c}{+32.4 [+30.7,+34.1]} \\
 & DominoTree (32) & \textbf{123.8} & \textbf{3.73} & \textbf{80.2} & \textbf{3.32} & 49.9 & \textbf{2.97} \\
 & \quad\emph{vs.\ Domino}, $\Delta$TPS\% & \multicolumn{2}{c}{+11.6 [+10.1,+13.2]} & \multicolumn{2}{c}{+5.2 [+3.5,+7.0]} & \multicolumn{2}{c}{-0.5 [-2.6,+1.5]$^{*}$} \\
 & \quad\emph{vs.\ Domino}, $\Delta\tau$\% & \multicolumn{2}{c}{+43.9 [+41.5,+46.4]} & \multicolumn{2}{c}{+41.8 [+39.3,+44.2]} & \multicolumn{2}{c}{+43.4 [+41.7,+45.2]} \\
\midrule
\multicolumn{8}{c}{\textbf{Qwen3.6-27B}} \\
\midrule
$\infty$Bench-Sum & AR & 64.9 & 1.00 & 60.8 & 1.00 & 53.5 & 1.00 \\
 & MTP & 119.9 & 2.62 & 106.1 & 2.59 & \textbf{84.5} & 2.58 \\
 & DFlash & 107.7 & 2.49 & 95.4 & 2.43 & 78.7 & 2.43 \\
 & Domino & 112.9 & 2.65 & 95.2 & 2.44 & 73.3 & 2.25 \\
 & DominoTree (15) & \textbf{135.9} & \textbf{3.47} & \textbf{113.3} & \textbf{3.24} & 84.0 & \textbf{2.97} \\
 & \quad\emph{vs.\ Domino}, $\Delta$TPS\% & \multicolumn{2}{c}{+20.2 [+18.4,+22.1]} & \multicolumn{2}{c}{+19.0 [+17.6,+20.5]} & \multicolumn{2}{c}{+14.4 [+12.7,+16.2]} \\
 & \quad\emph{vs.\ Domino}, $\Delta\tau$\% & \multicolumn{2}{c}{+31.1 [+29.5,+32.8]} & \multicolumn{2}{c}{+32.5 [+31.0,+34.0]} & \multicolumn{2}{c}{+32.0 [+30.7,+33.4]} \\
\cmidrule(lr){1-8}
Multi-LexSum & AR & 60.1 & 1.00 & 52.0 & 1.00 & 39.4 & 1.00 \\
 & MTP & 112.0 & 2.90 & 87.2 & 2.95 & \textbf{56.8} & 2.98 \\
 & DFlash & 108.7 & 2.98 & 85.0 & 2.98 & 56.6 & 3.01 \\
 & Domino & 117.0 & 3.30 & 86.8 & 3.13 & 55.2 & 2.88 \\
 & DominoTree (15) & \textbf{128.2} & \textbf{4.22} & \textbf{94.0} & \textbf{3.95} & 56.3 & \textbf{3.72} \\
 & \quad\emph{vs.\ Domino}, $\Delta$TPS\% & \multicolumn{2}{c}{+9.6 [+7.8,+11.2]} & \multicolumn{2}{c}{+8.3 [+7.2,+9.5]} & \multicolumn{2}{c}{+2.7 [+0.3,+4.8]} \\
 & \quad\emph{vs.\ Domino}, $\Delta\tau$\% & \multicolumn{2}{c}{+28.0 [+26.4,+29.7]} & \multicolumn{2}{c}{+26.2 [+24.4,+28.1]} & \multicolumn{2}{c}{+29.4 [+27.6,+31.1]} \\
\bottomrule
\end{tabular*}
\end{table*}

\begin{table*}[p]
\centering\footnotesize
\caption{Per-dataset SGLang concurrency, Qwen3-4B and Qwen3-8B, settings as in
Table~\ref{tab:sglang-conc}: goodput (tok/s) at each $c$, and $\tau$ averaged over $c$ (largest
spread within a row: $2.9\%$). GSM8K and MBPP use 128 prompts, MT-Bench 80. Bold marks the best
value per column within a dataset or within the overall rows.}
\label{tab:app-conc}
\setlength{\tabcolsep}{3pt}
\begin{tabular*}{\textwidth}{@{\extracolsep{\fill}}llcccccc}
\toprule
Dataset & Method & $c{=}2$ & $c{=}4$ & $c{=}8$ & $c{=}16$ & $c{=}32$ & $\tau$ \\
\midrule
\multicolumn{8}{c}{\textit{Qwen3-4B}} \\
\midrule
GSM8K & AR & 296 & 571 & 983 & 1758 & 3060 & -- \\
 & EAGLE-3 & 393 & 743 & 1107 & 1580 & 1935 & 3.35 \\
 & DFlash & 1059 & 1893 & 2820 & 3978 & 4940 & 6.46 \\
 & Domino & 1414 & \textbf{2314} & \textbf{3531} & \textbf{4750} & \textbf{5991} & 9.65 \\
 & DominoTree (32) & \textbf{1415} & 2303 & 2847 & 3397 & 4327 & \textbf{11.91} \\
\midrule
MBPP & AR & 295 & 570 & 991 & 1769 & 3076 & -- \\
 & EAGLE-3 & 388 & 667 & 1081 & 1609 & 1838 & 2.98 \\
 & DFlash & 969 & 1753 & \textbf{2733} & 3822 & 4685 & 6.06 \\
 & Domino & 978 & \textbf{1778} & 2728 & \textbf{3899} & \textbf{4900} & 6.91 \\
 & DominoTree (32) & \textbf{1078} & 1763 & 2206 & 2771 & 3533 & \textbf{9.09} \\
\midrule
MT-Bench & AR & 294 & 563 & 943 & 1689 & \textbf{2895} & -- \\
 & EAGLE-3 & 389 & 658 & 1064 & 1569 & 1807 & 2.94 \\
 & DFlash & 569 & 1017 & 1540 & 2153 & 2511 & 4.10 \\
 & Domino & 613 & 1100 & \textbf{1697} & \textbf{2369} & 2816 & 5.05 \\
 & DominoTree (32) & \textbf{698} & \textbf{1138} & 1370 & 1757 & 2189 & \textbf{6.61} \\
\midrule\midrule
\multicolumn{8}{c}{\textit{Qwen3-4B -- Overall (mean of the three datasets above)}} \\
\midrule
Overall & AR & 295 & 568 & 972 & 1739 & 3011 & -- \\
 & EAGLE-3 & 390 & 689 & 1084 & 1586 & 1860 & 3.09 \\
 & DFlash & 866 & 1555 & 2364 & 3318 & 4045 & 5.54 \\
 & Domino & 1002 & 1731 & \textbf{2652} & \textbf{3672} & \textbf{4569} & 7.21 \\
 & DominoTree (32) & \textbf{1063} & \textbf{1734} & 2141 & 2642 & 3350 & \textbf{9.20} \\
\midrule
\multicolumn{8}{c}{\textit{Qwen3-8B}} \\
\midrule
GSM8K & AR & 185 & 350 & 665 & 1221 & 1732 & -- \\
 & EAGLE-3 & 273 & 511 & 807 & 1094 & 1363 & 3.30 \\
 & DFlash & 596 & 1134 & 1770 & 2227 & 2681 & 6.54 \\
 & Domino & 837 & \textbf{1546} & \textbf{2373} & \textbf{3025} & \textbf{3767} & 9.76 \\
 & DominoTree (32) & \textbf{859} & 1388 & 1801 & 2269 & 2713 & \textbf{11.95} \\
\midrule
MBPP & AR & 185 & 351 & 663 & 1204 & 1701 & -- \\
 & EAGLE-3 & 241 & 451 & 715 & 953 & 1179 & 2.87 \\
 & DFlash & 539 & 1028 & 1634 & 2065 & 2606 & 5.94 \\
 & Domino & 604 & \textbf{1136} & \textbf{1794} & \textbf{2253} & \textbf{2786} & 6.94 \\
 & DominoTree (32) & \textbf{666} & 1096 & 1437 & 1808 & 2218 & \textbf{9.17} \\
\midrule
MT-Bench & AR & 185 & 344 & 628 & 1139 & 1566 & -- \\
 & EAGLE-3 & 228 & 417 & 650 & 885 & 1068 & 2.74 \\
 & DFlash & 319 & 599 & 943 & 1167 & 1419 & 4.13 \\
 & Domino & 370 & 689 & \textbf{1085} & \textbf{1360} & \textbf{1706} & 5.12 \\
 & DominoTree (32) & \textbf{433} & \textbf{706} & 909 & 1148 & 1367 & \textbf{6.79} \\
\midrule\midrule
\multicolumn{8}{c}{\textit{Qwen3-8B -- Overall (mean of the three datasets above)}} \\
\midrule
Overall & AR & 185 & 348 & 652 & 1188 & 1667 & -- \\
 & EAGLE-3 & 247 & 459 & 724 & 977 & 1204 & 2.97 \\
 & DFlash & 484 & 920 & 1449 & 1820 & 2236 & 5.54 \\
 & Domino & 604 & \textbf{1123} & \textbf{1751} & \textbf{2212} & \textbf{2753} & 7.27 \\
 & DominoTree (32) & \textbf{653} & 1064 & 1382 & 1742 & 2099 & \textbf{9.30} \\
\bottomrule
\end{tabular*}
\end{table*}

\begin{table*}[p]
\centering\footnotesize
\caption{As Table~\ref{tab:app-conc}, for Qwen3.6-27B (one H200, tree budget 15). \textbf{MTP}
takes EAGLE-3's place, since no EAGLE-3 checkpoint exists for this model.}
\label{tab:app-conc-27b}
\setlength{\tabcolsep}{3pt}
\begin{tabular*}{\textwidth}{@{\extracolsep{\fill}}llcccccc}
\toprule
Dataset & Method & $c{=}2$ & $c{=}4$ & $c{=}8$ & $c{=}16$ & $c{=}32$ & $\tau$ \\
\midrule
\multicolumn{8}{c}{\textit{Qwen3.6-27B}} \\
\midrule
GSM8K & AR & 127 & 238 & 426 & 339 & 606 & -- \\
 & MTP & 323 & 572 & 937 & 862 & 1095 & 3.67 \\
 & DFlash & 552 & 901 & 1290 & 1171 & 1121 & 7.40 \\
 & Domino & 673 & \textbf{1068} & \textbf{1532} & \textbf{1466} & \textbf{1456} & 9.31 \\
 & DominoTree (15) & \textbf{674} & 1058 & 1468 & 1373 & 1378 & \textbf{9.78} \\
\midrule
MBPP & AR & 125 & 227 & 384 & 311 & 514 & -- \\
 & MTP & 295 & 498 & 774 & 745 & 916 & 3.63 \\
 & DFlash & 476 & 744 & 1036 & 1054 & 1050 & 7.66 \\
 & Domino & 480 & 757 & \textbf{1083} & \textbf{1078} & \textbf{1091} & 7.51 \\
 & DominoTree (15) & \textbf{500} & \textbf{775} & 1045 & 1063 & 1060 & \textbf{8.08} \\
\midrule
MT-Bench & AR & 127 & 236 & 420 & 335 & 583 & -- \\
 & MTP & 272 & 484 & 794 & 745 & \textbf{926} & 3.16 \\
 & DFlash & 306 & 515 & 763 & 638 & 638 & 4.86 \\
 & Domino & 334 & 559 & 833 & 731 & 727 & 5.41 \\
 & DominoTree (15) & \textbf{388} & \textbf{623} & \textbf{890} & \textbf{784} & 767 & \textbf{6.15} \\
\midrule\midrule
\multicolumn{8}{c}{\textit{Qwen3.6-27B -- Overall (mean of the three datasets above)}} \\
\midrule
Overall & AR & 126 & 234 & 410 & 328 & 568 & -- \\
 & MTP & 296 & 518 & 835 & 784 & 979 & 3.49 \\
 & DFlash & 444 & 720 & 1030 & 954 & 936 & 6.64 \\
 & Domino & 496 & 794 & \textbf{1149} & \textbf{1092} & \textbf{1091} & 7.41 \\
 & DominoTree (15) & \textbf{521} & \textbf{819} & 1134 & 1073 & 1068 & \textbf{8.01} \\
\bottomrule
\end{tabular*}
\end{table*}

\begin{table*}[p]
\centering\footnotesize
\caption{Tuning DominoTree's tree budget and candidate width on Qwen3.6-27B (one H200), 20 tuning
prompts. Throughput is completion tokens over wall time, prefill included; $\tau$ is mean
per-prompt accepted length. Every row was checked lossless against the target's own decoding
(Section~\ref{sec:exp-lossless}). Bold marks the best throughput and, separately, the best
$\tau$ over the budget sweep and within each $k$ block of the width sweep; the timing panel is
not bolded.}
\label{tab:app-qwen27b-ablation}

\begin{center}
\begin{minipage}[t]{0.30\textwidth}
\centering
\textbf{Node-budget sweep}\\[1pt]
{\footnotesize (candidate width $M{=}128$, branching factor $k{=}8$)}\\[5pt]
\begin{tabular}{rrr}
\toprule
Budget $B$ & Tok/s & $\tau$ \\
\midrule
8  & 269.9 & 6.70 \\
12 & 313.1 & 8.57 \\
15 & \textbf{345.5} & 9.61 \\
32 & 316.6 & 10.59 \\
64 & 273.9 & \textbf{11.17} \\
\bottomrule
\end{tabular}
\end{minipage}%
\hfill
\begin{minipage}[t]{0.66\textwidth}
\centering
\textbf{Per-step timing}\\[1pt]
{\footnotesize (mean milliseconds per decoding step for each pipeline stage, averaged over the
run, at $k{=}8$, $M{=}128$; totals sum the unrounded per-stage means)}\\[5pt]
\begin{tabular}{rrrrrrrr}
\toprule
$B$ & Draft & Build & Prepare & Verify & Accept & Commit & Total \\
\midrule
8  & 1.740 & 1.384 & 0.585 & 17.056 & 0.251 & 0.226 & 21.241 \\
15 & 1.742 & 1.561 & 0.629 & 18.056 & 0.282 & 0.275 & 22.544 \\
32 & 1.869 & 2.964 & 0.591 & 22.068 & 0.228 & 0.261 & 27.981 \\
64 & 1.872 & 3.082 & 0.592 & 29.513 & 0.249 & 0.260 & 35.568 \\
\bottomrule
\end{tabular}
\end{minipage}
\end{center}

\bigskip
\begin{center}
\textbf{Candidate-width and branching-factor sweep}, at the fixed budget $B{=}15$
($M{=}0$ means correcting over the full vocabulary instead of a restricted candidate
list). This panel was measured in an earlier session than the budget sweep; its
$(k,M){=}(8,128)$ setting and the sweep's $B{=}15$ row measure the same configuration and differ by
$1\%$ in throughput.
\end{center}

\begin{center}
\begin{minipage}[t]{0.31\textwidth}
\centering
$k{=}4$\\[3pt]
\begin{tabular}{rrr}
\toprule
$M$ & Tok/s & $\tau$ \\
\midrule
32  & 334.8 & 9.57 \\
64  & 338.6 & \textbf{9.61} \\
128 & \textbf{342.0} & \textbf{9.61} \\
0   & 248.6 & 9.59 \\
\bottomrule
\end{tabular}
\end{minipage}%
\hfill
\begin{minipage}[t]{0.31\textwidth}
\centering
$k{=}8$\\[3pt]
\begin{tabular}{rrr}
\toprule
$M$ & Tok/s & $\tau$ \\
\midrule
32  & 335.9 & 9.58 \\
64  & 338.8 & \textbf{9.61} \\
128 & \textbf{342.2} & \textbf{9.61} \\
0   & 257.3 & \textbf{9.61} \\
\bottomrule
\end{tabular}
\end{minipage}%
\hfill
\begin{minipage}[t]{0.31\textwidth}
\centering
$k{=}16$\\[3pt]
\begin{tabular}{rrr}
\toprule
$M$ & Tok/s & $\tau$ \\
\midrule
32  & 337.4 & 9.58 \\
64  & \textbf{341.6} & \textbf{9.61} \\
128 & 339.9 & \textbf{9.61} \\
0   & 253.7 & 9.59 \\
\bottomrule
\end{tabular}
\end{minipage}
\end{center}

\end{table*}

\end{document}